\documentclass[letterpaper]{llncs}
\usepackage[T1]{fontenc}
\usepackage{graphicx}
\usepackage{booktabs}
\usepackage{cite}
\usepackage{xcolor}
\usepackage{adjustbox}
\usepackage{multirow}
\usepackage{tabularx}
\usepackage{caption}
\usepackage[position=bottom]{subcaption}
\usepackage{amsmath}
\usepackage{makecell} 
\usepackage{lipsum}
\usepackage{setspace} 

\usepackage{amssymb}
\usepackage{hyperref}
\usepackage{algorithm}
\usepackage{algpseudocode}

\usepackage{framed}
\makeatletter

\begin{document}
\title{Distribution-Aware Robust Learning from Long-Tailed Data with Noisy Labels}
\titlerunning{DaSC}
%
\author{Jae Soon Baik\inst{1} \and
In Young Yoon\inst{1} \and Kun Hoon Kim\inst{1} \and Jun Won Choi\inst{2}}
\authorrunning{J.S. Baik et al.}
%
\institute{Hanyang University, Korea \\
\email{\{jsbaik, inyoungyoon, khkim\}@spa.hanyang.ac.kr,}
\and 
Seoul National University, Korea \\
\email{junwchoi@snu.ac.kr}}
\maketitle

\begin{abstract}
Deep neural networks have demonstrated remarkable advancements in various fields using large, well-annotated datasets. However, real-world data often exhibit long-tailed distributions and label noise, significantly degrading generalization performance. Recent studies addressing these issues have focused on noisy sample selection methods that estimate the centroid of each class based on high-confidence samples within each target class. The performance of these methods is limited because they use only the training samples within each class for class centroid estimation, making the quality of centroids susceptible to long-tailed distributions and noisy labels.
In this study, we present a robust training framework called \textit{Distribution-aware Sample Selection and Contrastive Learning} (DaSC). Specifically, DaSC introduces a Distribution-aware Class Centroid Estimation (DaCC) to generate enhanced class centroids. DaCC performs weighted averaging of the features from all samples, with weights determined based on model predictions. Additionally, we propose a confidence-aware contrastive learning strategy to obtain balanced and robust representations. The training samples are categorized into high-confidence and low-confidence samples. Our method then applies Semi-supervised Balanced Contrastive Loss (SBCL) using high-confidence samples, leveraging reliable label information to mitigate class bias. For the low-confidence samples, our method computes Mixup-enhanced Instance Discrimination Loss (MIDL) to improve their representations in a self-supervised manner. Our experimental results on CIFAR and real-world noisy-label datasets demonstrate the superior performance of the proposed DaSC compared to previous approaches. Our code is available at \href{https://github.com/JaesoonBaik1213/DaSC}{https://github.com/JaesoonBaik1213/DaSC}.

\keywords{Noisy label \and Long-tailed learning \and Contrastive learning.}
\end{abstract}

\begin{figure*}[t]
    \centering
    \begin{subfigure}[b]{0.27\textwidth}
        \centering
        \includegraphics[width=0.995\textwidth]{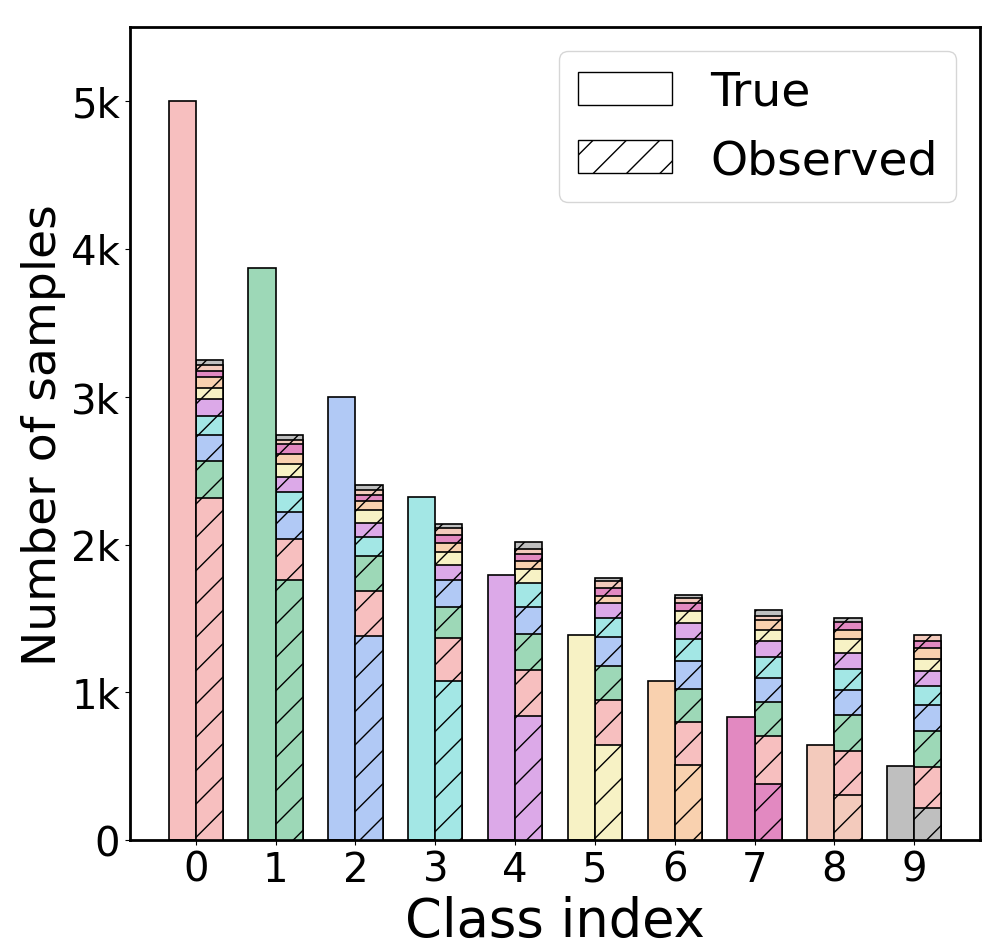}
        \caption[]{Class Distribution}
        \label{fig: class distribution on sym}
    \end{subfigure}
    \hfill
    \begin{subfigure}[b]{0.33\textwidth}
        \centering 
        \includegraphics[width=1.0\textwidth]{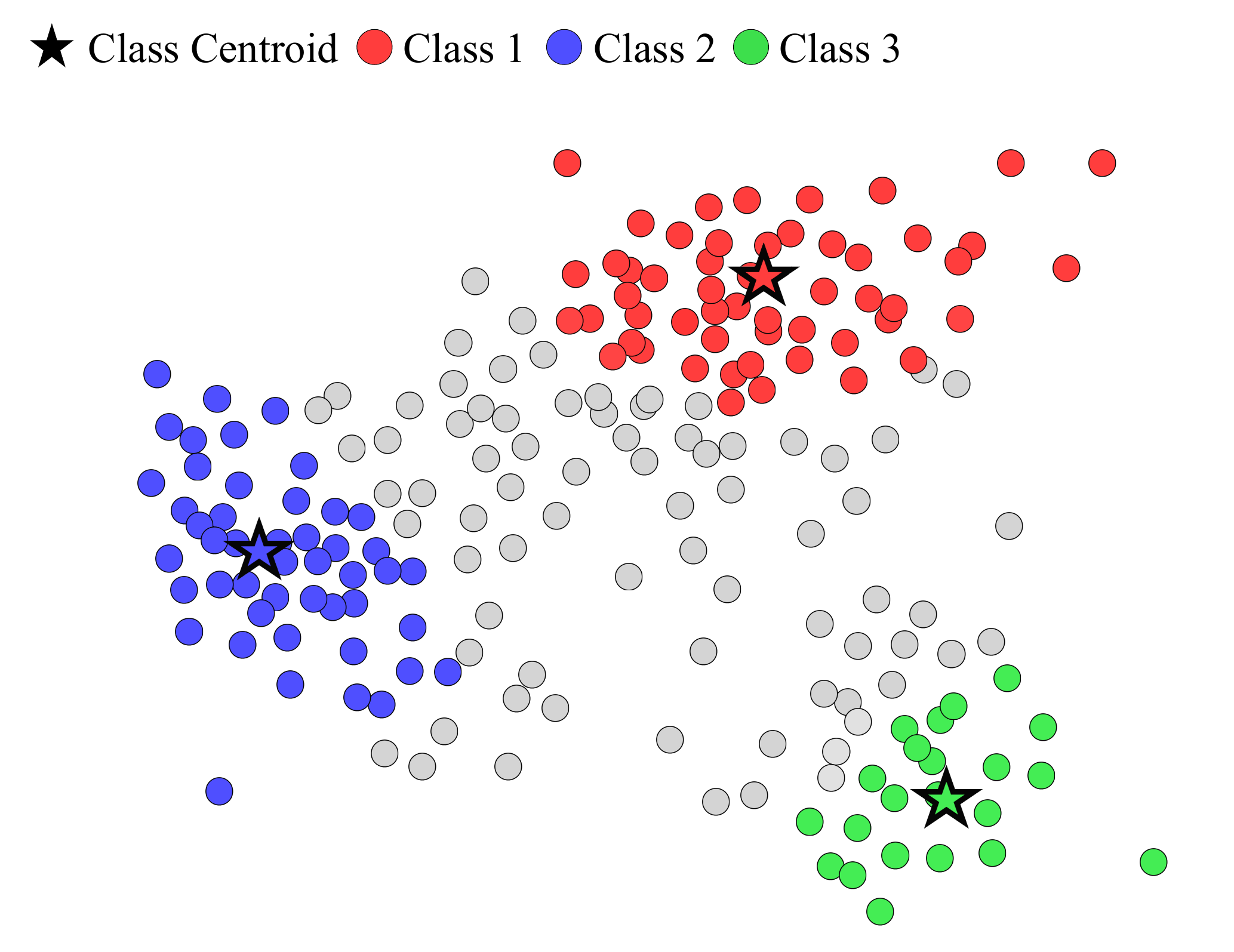}
        \caption[]{Previous method}
        \label{fig: previous centroid estimation}
    \end{subfigure}
    \hfill
    \begin{subfigure}[b]{0.33\textwidth}
        \centering
        \includegraphics[width=1.0\textwidth]{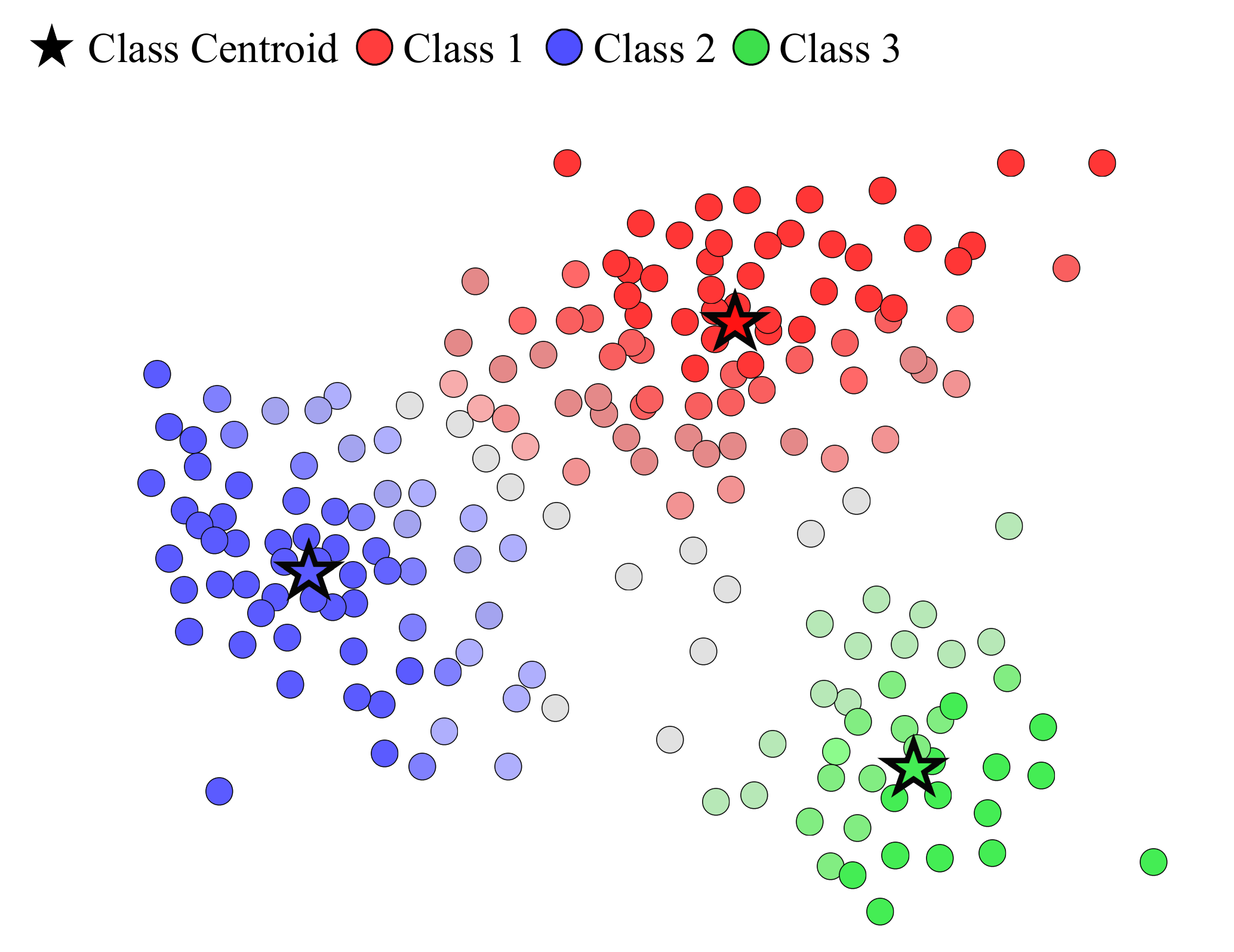}
        \caption[]{Proposed method}
        \label{fig: our centroid estimation}
    \end{subfigure}
    \hfill

    \captionsetup{singlelinecheck = false}
    \caption[]{(a) Class distribution on long-tailed CIFAR-10 with symmetric noise of 0.6. (b) and (c) show the illustration of previous class centroid estimation and our distribution-aware class centroid estimation, respectively. In the previous methods, samples with low confidence, depicted as grey dots, are excluded from the class centroid estimation. In contrast, our approach leverages all samples for centroid estimation, using their prediction values as weights. The colors of the samples indicate the prediction values.}
    \label{fig: class distribution and sample selection}
\end{figure*}

\section{Introduction}
\label{sec: intro}
In recent years, notable achievements have been made in various machine-learning applications owing to large-scale datasets with high-quality annotated labels \cite{dosovitskiy2020image,he2016deep}. However, in practice, obtaining a large-scale dataset with accurately annotated labels is challenging and expensive. A realistic alternative is to collect data from the Internet through web crawling, but this is typically affected by long-tail distributions as well as mislabeled data \cite{xiao2015learning}. Training a model on such datasets results in biases toward head classes and suffers from memorization of label noise \cite{bo2018coteaching}, consequently leading to poor generalization performance. 

Numerous methods have been explored to address the challenges of long-tailed distributions and noisy labels. In long-tailed (LT) learning, re-balancing techniques are the most straightforward method to mitigate bias toward head classes. These techniques include oversampling, undersampling \cite{zhou2020bbn,chawla2002smote}, and adjusting instance-level or class-level weights in loss functions \cite{cao2019ldam,cui2019class,ren2018learning,shu2019metaweight}. Recently, inspired by the success of supervised contrastive learning (SCL) \cite{khosla2020supervised}, several studies \cite{zhu2022balanced,wang2021hybridsc,kang2020balfeat,cui2021parametric,li2022targeted} have aimed to generate balanced representations by incorporating class distribution information into SCL. For learning with noisy labels (NL), various strategies have been proposed. These include robust loss functions \cite{ren2018learning,liu2020early}, sample selection methods for distinguishing clean from noisy samples \cite{bo2018coteaching,arazo2019unsupervised,tanaka2018joint,huang2023twin}, and correcting noisy labels using model predictions \cite{li2020dividemix,tanaka2018joint,reed2014training}. Semi-supervised learning (SSL) has also been applied to enhance learning performance by utilizing noisy data in training \cite{li2020dividemix,karim2022unicon}. These methods often employ suitable distance metrics to identify noisy samples, which are then utilized as unlabeled data for SSL. Recent studies have leveraged SCL with reliable label information \cite{li2022selective,huang2023twin,ortego2021multi,zheltonozhskii2022contrast} and unsupervised contrastive learning \cite{chen2020simple,wu2018unsupervisedIDC} to produce representations robust to noisy labels.

Despite the success of individual approaches, simultaneously addressing long-tailed distributions and label noise remains a significant challenge. The combined issue of noisy labels and long-tailed distributions (NL-LT) complicates the accurate identification of true data distributions, significantly degrading the effectiveness of these methods. Recent studies \cite{wei2021robust,wei2022prototypical,li2023sfa,lu2023tabasco} have employed feature-based noisy sample selection methods to identify noisy labels. As illustrated in Fig. \ref{fig: previous centroid estimation}, these methods rely on the distance of input features from the class centroids, which are obtained by averaging the features of high-confidence samples within each class.

However, this approach has several limitations. First, it uses only a portion of the training data within each class for estimating class centroids. As shown in Fig. \ref{fig: class distribution on sym}, samples mislabeled as other classes are excluded from the centroid estimation process. This exclusion results in insufficient data samples to reliably determine class centroids, which is particularly problematic for tail classes.
Second, this method assigns equal weight to all samples when calculating class centroids. In the presence of labeling errors, this uniform weighting overlooks the potential for certain samples to be erroneous.
Last, this approach does not actively seek to enhance representation quality. This limitation diminishes its effectiveness for tail classes, where obtaining high-quality representations is more critical than for head classes.

We aim to address these limitations by introducing a novel feature-based noisy sample selection method called Distribution-aware Class Centroid Estimation (DaCC). As shown in Fig. \ref{fig: our centroid estimation}, DaCC estimates class centroids by weighted averaging the features of samples from various classes. DaCC determines these weights based on model predictions, without relying on noisy labels, allowing for the adaptive inclusion of data samples from different classes in the centroid estimation process. Consequently, our method effectively utilizes informative samples with high-confidence scores when estimating class centroids.

Furthermore, we propose a Confidence-aware Contrastive Learning strategy to achieve balanced and robust representations in noisy, long-tailed classification. Specifically, we categorize samples into high-confidence and low-confidence groups based on the confidence scores derived from their pseudo-labels. For the high-confidence group, we employ Semi-supervised Balanced Contrastive Loss (SBCL), which mitigates class bias using confident label information. Simultaneously, for the low-confidence group, we introduce Mixup-enhanced Instance Discrimination Loss (MIDL), which enhances data representation in a self-supervised manner. MIDL utilizes mixup augmentation \cite{Zhang2018mixup} to construct diverse negative keys, which leads to more robust representations for the low-confidence group.

We integrate all the aforementioned components, presenting a novel training framework, referred to as \textit{Distribution-aware Sample Selection and Contrastive Learning (DaSC)}. We conduct experiments on both modified CIFAR datasets and real-world datasets that exhibit NL-LT. The experiments demonstrate that DaSC achieves significant performance improvements over existing methods.

The key contributions of this study are summarized as follows:
\begin{itemize}
    \item We present a novel class centroid estimation method called DaCC. This method utilizes the entire set of training samples as compared to the existing methods. Class centroids are generated by averaging samples with weights obtained by temperature-scaled model predictions. This strategy leads to more reliable class centroid estimation and efficient use of training data.
    \item We also introduce a confidence-aware contrastive learning strategy to improve the quality of feature representations. This method applies two different contrastive losses, SBCL and MIDL, to low-confidence and high-confidence sample groups, respectively. SBCL uses label information to achieve a balanced representation of high-confidence samples, while MIDL employs a mixup augmentation to learn a robust representation for low-confidence samples.
    \item  Our experimental results confirm that DaSC achieves state-of-the-art performance across various configurations.
\end{itemize}

\section{Related Work}
\label{sec: related work}

\subsection{Long-Tailed Learning}
\label{sec: long-tailed learning}
Numerous approaches have been proposed to address the challenges of learning from long-tailed data. Re-balancing is a classical strategy for addressing class bias, including data re-sampling \cite{chawla2002smote,shen2016relay} and loss re-weighting \cite{cao2019ldam,cui2019class,menon2021logitadjust,shu2019metaweight}. Recently, long-tailed data issues have been studied from the perspectives of both representation learning and classifier learning \cite{kang2020decouple,menon2021logitadjust}. Ensemble-based methods have also been proposed to achieve robust performance under long-tailed data distributions \cite{wang2021ride,zhou2020bbn}.

\subsection{Learning with Noisy Labels}
\label{sec: learning with noisy labels}
Various methods have been explored to achieve robust learning against noisy labels. These methods can be broadly classified into three approaches: 1) estimation of the noise transition matrix \cite{patrini2017fcorrection,cheng2022class,xia2020part,yao2020dual}; 2) robust loss functions \cite{ren2018learning,wang2017multiclass,liu2020early}; and 3) noisy sample selection \cite{bo2018coteaching,arazo2019unsupervised,tanaka2018joint,jiang2018mentornet,li2020dividemix,huang2023twin}. Recently, some studies have explored applying SSL to address a noisy label issue \cite{li2020dividemix,karim2022unicon,zhao2022centrality}. These methods treated clean samples as labeled data and samples with noisy labels as unlabeled data after detecting correctly labeled samples.

\subsection{Long-Tailed Learning with Noisy Labels}
\label{sec: long-tailed learning with noisy labels}
A series of studies have been conducted to resolve both noisy labels and long-tailed distributions. Meta-learning methods \cite{ren2018learning,shu2019metaweight} have been proposed to prioritize clean data by adaptively learning explicit instance weights from metadata. HAR \cite{cao2021har} introduced an adaptive Lipschitz regularization that takes into account the uncertainty inherent in a particular data sample. To better differentiate mislabeled examples, RoLT \cite{wei2021robust} and PCL \cite{wei2022prototypical} calculated a centroid of each class by taking the average of the features with high confidence in the corresponding target class. The obtained class centroids were subsequently used for noisy sample selection in RoLT and as part of the classifier in PCL. Recently, SFA \cite{li2023sfa} adopted a Gaussian distribution to estimate the distribution of class centroids, addressing uncertainty in the representation space. TABASCO \cite{lu2023tabasco} employed two different metrics, one derived from model predictions and the other from the estimated class centroids. It then utilized a Gaussian Mixture Model (GMM) to determine the most appropriate metric for detecting correctly labeled samples with different noise types. 

\subsection{Contrastive Learning for Addressing Long-Tailed Distribution and Noisy Labels}
Contrastive learning aims to enhance the representation by comparing sample pairs, pulling semantically similar samples closer together while pushing semantically dissimilar samples further apart. Contrastive learning has been successfully applied to various tasks \cite{he2020momentum,chen2020simple,han2020videorep,wang2022cross}. Recently, SCL \cite{khosla2020supervised} introduced fully supervised contrastive learning to encourage the samples of the same class to lie closer in feature space compared to those of different classes. Several LT methods \cite{zhu2022balanced,wang2021hybridsc,kang2020balfeat,cui2021parametric,li2022targeted} have enhanced SCL by utlizing supervision provided by long-tailed training data. PaCo \cite{cui2021parametric} introduced parametric learnable class centers to learn a balanced representation. BCL \cite{zhu2022balanced} proposed a balanced contrastive loss that uses logit compensation to induce all samples to form a regular simplex. In NL learning, several methods \cite{li2022selective,huang2023twin,ortego2021multi,zheltonozhskii2022contrast} have employed contrastive learning \cite{chen2020simple,he2020momentum,khosla2020supervised} to better distinguish between noisy and clean data. MOIT \cite{ortego2021multi} proposed interpolated contrastive learning that combines mixup with SCL to learn robust representations. Sel-CL \cite{li2022selective} identified confidence pairs and modeled their pair-wise relations to obtain a robust representation.

\section{Proposed Method}
\label{sec: Proposed Method}

\subsection{Overview of the Proposed Method}
\label{sec: overview of the proposed method}
{\bf Problem Setup.} Consider a $K$-class classification task. We train the model using the training dataset $\mathcal{D}=\{(x(i), \tilde{y}(i))\}^N_{i=1}$, where $N$ denotes the total number of training samples, $x(i)$ denotes the $i$-th training sample, and $\tilde{y}(i) \in \{1, ...,K\}=\mathcal{Y}$ is its associated label. As a result of labeling errors, a fraction of the training samples are incorrectly labeled, i.e., $\exists i\in [1, N], \tilde{y}(i) \neq y^*(i)$, where $y^*(i)$ indicates the true label for the sample $x(i)$. We denote the training sample set of class $k$ as $\mathcal{D}_k = \{(x(i), 
\tilde{y}(i))|\tilde{y}(i)=k\}$. The training dataset $\mathcal{D}$ follows a long-tailed distribution with an imbalance ratio of $\rho$, where $\rho=\min_{k}|\mathcal{D}_k|/\max_{k}|\mathcal{D}_k|$. Our goal is to achieve robust classification performance on unseen data, given a long-tailed noisy training dataset $\mathcal{D}$.

\begin{figure*}[t]
    \begin{center}
    \includegraphics[width=1\textwidth]{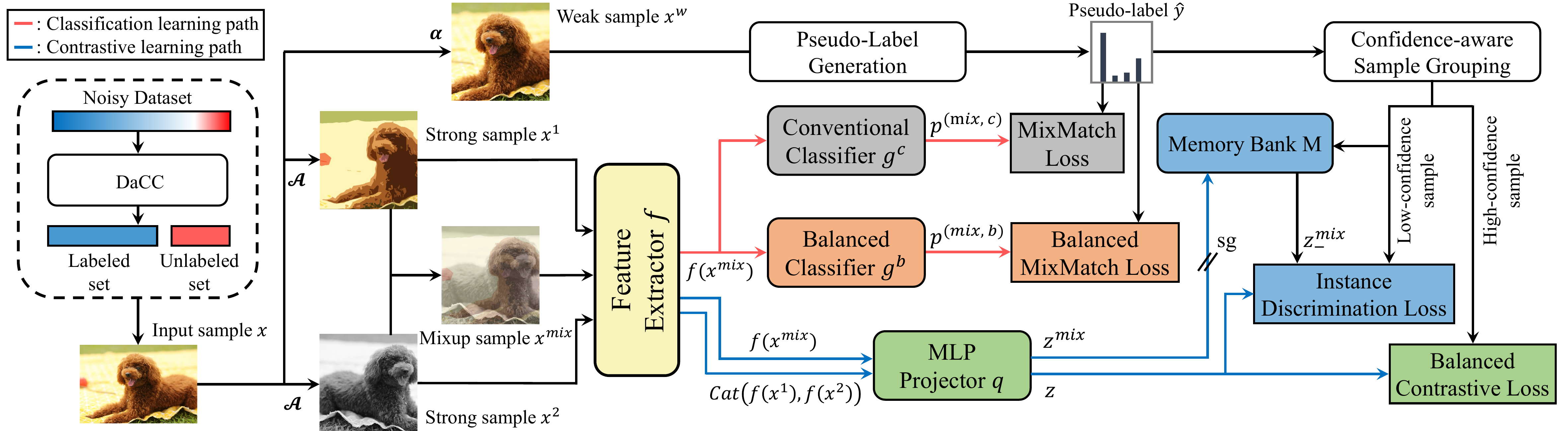}
    \end{center}
    \caption{{\bf Illustration of the proposed DaSC framework:} Given a dataset, the DaSC detects correctly labeled samples using class centroids obtained from DaCC. Subsequently, the model is trained under a semi-supervised learning (SSL) framework, where confidence-aware contrastive learning is employed to enhance the representations. $\texttt{Cat}$ and $\texttt{sg}$ denotes concatenation and stop-gradient operations, respectively.}
    \label{fig: Overview of proposed method}
\end{figure*}

\noindent {\bf Overview.} Fig. \ref{fig: Overview of proposed method} illustrates the overview of the proposed DaSC framework. Our architecture comprises DaCC, a sample selection module, a shared feature extractor $f$, a conventional classifier $g^c$, a balanced classifier $g^b$, a pseudo-label generator, and a Multi-Layer Perceptron (MLP) projector $q$. Initially, DaCC computes the class centroids. Then, correctly labeled samples are identified based on their distance from the class centroids. The next step is to train the model using noisy samples as unlabeled samples within the SSL framework \cite{berthelot2019mixmatch}. For each training sample $(x(i), \tilde{y}(i))$, two strongly augmented samples $x^1(i)$ and $x^2(i)$ and a weakly augmented sample $x^w(i)$ are generated \cite{li2020dividemix}. Additionally, a mixup sample $(x^{mix}(i), y^{mix}(i))$ is created from $x^1(i)$ and $x^2(i)$ using mixup augmentation \cite{Zhang2018mixup}.
The conventional classifier $g^c$ is applied to $x^{mix}(i)$ to produce a model prediction $p^{(mix, c)}(i) = \left[p^{(mix, c)}_{1}(i), \ldots, p^{(mix, c)}_{K}(i)\right]^T$, where $p^{(mix, c)}_{k}(i)$ denotes the predicted probability of the $k$-th class. The balanced classifier $g^b$ is then applied to $x^{mix}(i)$ to produce a model prediction $p^{(mix, b)}(i)$. This balanced classifier uses Balanced Softmax \cite{ren2020balsoftmax} to mitigate class bias. Following the DivideMix \cite{li2020dividemix}, we employ a pseudo-label generator to produce a pseudo-label $\hat{y}(i)$, based on the provided labels and model predictions for labeled samples or solely on the model predictions for unlabeled samples. 
The projector $q$ is applied to both $x^{mix}(i)$ and the concatenation of $x^1(i)$ and $x^2(i)$, producing lower-dimensional representations $z^{mix}(i)$ and $z(i)$, respectively. These representations are used in confidence-aware contrastive learning to enhance representation quality. Specifically, we classify an input sample $x(i)$ into a high-confidence sample or a low-confidence sample by comparing the maximum confidence score in $\hat{y}(i)$ with a threshold $\tau_c$. Subsequently, we employ SBCL for the high-confidence sample group and MIDL for the low-confidence sample group.
 
\subsection{Distribution-Aware Class Centroid Estimation}
\label{sec: distribution-aware class centroid estimation}
Conventional sample selection methods proposed in \cite{li2023sfa,wei2021robust,lu2023tabasco} used only target class samples to estimate their class centroids.  In contrast, the proposed DaCC method estimates class centroids using samples from all classes, weighting each sample based on its confidence score. However, noisy labels can adversely affect the class centroids. To mitigate this, we apply temperature scaling (TS) to the model predictions, reducing the weights for unreliable samples and increasing them for more reliable ones.

DaCC takes the $i$-th image $x(i)$ as input and produces the low-dimensional features $z'(i)$ using the backbone network followed by the MLP projector. Then, it applies the conventional classifier and the balanced classifier to produce the prediction outputs $p^c(i)=[p_1^c(i),...,p_K^c(i)]$ and $p^b(i)=[p_1^b(i),...,p_K^b(i)]$, respectively, where $k$ is the class index.  The class centroid $c_k$ for the $k$-th class is calculated as
\begin{align}
    c_k &= Norm\left(\sum_{x(i)\in\mathcal{D}^{I}}\hat{p}_{k}^c(i)z'(i)\right), 
    \end{align} 
where 
\begin{align}
    \hat{p}_{k}^c(i) &= \frac{\exp\left(p_{k}^c(i)/\tau_T\right)}{\sum_{k=1}^K \exp\left(p_{k}^c(i)/\tau_T\right)}, \\ 
    \mathcal{D}^{I} & = \{ x(i) | i \in [1,N], \max_{k} p^{c}_{k}(i) >\tau, \max_{k} p^{b}_{k}(i) >\tau \},
    \label{eq:centroid estimation}
\end{align}
where $\tau_T$ denotes the temperature parameter, $\tau$ denotes the confidence threshold, and $Norm$ denotes $l_2$ normalization. As training goes on, we gradually increase the confidence threshold as $\tau=\phi^t\hat{\tau}$, where $t$ is the epoch index. Following the setting from the previous works \cite{li2023sfa,wei2022prototypical,wei2021robust}, $\phi$ is set to $1.005$ and $\hat{\tau}$ is set to $1/K$. Note that the samples used to obtain the centroid for the $k$-th class belong to any classes.

After calculating the class centroids, we identify correctly labeled samples.  The assignment probability of a sample $x(i)$ belonging to class centroid $c_k$ is given by  
\begin{align}
    \Gamma(y(i)=k|z'(i), c) &= \frac{\exp(z'(i) \cdot c_k)}{\sum_{k} \exp(z'(i) \cdot c_{k})},
    \label{eq: centroid assignment}
\end{align}
where $x \cdot y$ denotes the dot-product operation. Subsequently, we fit a two-component GMM to maximize the likelihood of the assignment probability, i.e., $\Gamma(y(i)=k|x(i), c) \sim \sum^2_{j=1}w_{k,j} \mathcal{N}(\mu_{k,j}, \sigma_{k,j})$, where $\mu_{k,j}$ and $\sigma_{k,j}$ are the mean and variance of the $j$-th Gaussian component. We expect that correctly labeled samples will cluster close to the class centroids, while noisy samples will be more dispersed. Therefore, samples with a posterior probability exceeding 0.5 are classified as clean, and the remaining samples are identified as noisy.

\subsection{Confidence-Aware Contrastive Learning}
\label{sec: confidence-aware contrastive learning}
Confidence-aware contrastive learning applies different strategies to high-confidence and low-confidence samples. We classify a sample $x(i)$ by comparing the maximum confidence score in the pseudo-label $\hat{y}(i)$ with a threshold $\tau_c$. If the score exceeds the threshold, the sample is classified as a high-confidence one; otherwise, it is classified as a low-confidence one.

\noindent {\bf Semi-Supervised Balanced Contrastive Loss.} 
Given a set of high-confidence samples $B^{high}$ within a mini-batch $B$, the SBCL is expressed as 
\begin{align}
\mathcal{L}_{SBCL} = \frac{1}{|B^{high}|} \sum_{i \in B^{high}} \mathcal{L}_{SBCL}(i),
\end{align}
where
\begin{align}
    \mathcal{L}_{SBCL}(i) &= -\frac{1}{|B^{high}_{y_i}|} \sum_{p\in B^{high}_{y_i}\setminus\{i\}}\log \frac{\exp(z(i)\cdot z(p)/\tau_s)}{\sum_{j=1}^{K} \frac{1}{|B^{high}_j|}\sum_{k\in B^{high}_j} \exp(z(i)\cdot z(k)/\tau_s)},
    \label{eq: balanced contrastive loss}
\end{align}
where $A\setminus x$ denotes the set subtraction operation,  $y_i$ is the index of the largest entry of the pseudo-label $\hat{y}(i)$, $B^{high}_j$ represents the set of sample indices for high-confidence samples whose predictions correspond to the class $j$ within a mini-batch $B$, and $\tau_s$ is the temperature parameter. This method effectively leverages reliable label information from high-confidence samples to penalize head classes, aiding in the learning of balanced representations.

\noindent {\bf Mixup Enhanced Instance Discrimination Loss.} 
Building on a simple instance discrimination task \cite{wu2018unsupervisedIDC}, we propose mixup-enhanced instance discrimination loss for low-confidence samples. For each sample $x(i)$ within the mini-batch $B^{low}$ of low-confidence samples, we generate two augmented images, $x^1(i)$ and $x^2(i)$, serving as the {\it query} and {\it positive key}, respectively. We obtain the feature representations $z^1(i)$ and $z^2(i)$ from $x^1(i)$ and $x^2(i)$ by applying the backbone network, followed by the MLP projector. Additionally, we devise a variant that generates a {\it negative key} from a mixup image $x^{mix}$ \cite{Zhang2018mixup}. Note that mixup images are used to enhance the diversity of negative samples. The mixup image is obtained by taking a convex combination between $x^1(i)$ and $x^2(i)$ as
\begin{align}
    x^{mix}(i) &= \lambda x^1(i) + (1-\lambda) x^2(i),
    \label{eq: mixup}
\end{align}
where $\lambda$ denotes the mixing ratio obtained from Beta distribution, e.g., $\lambda\sim Beta(\alpha,\alpha)$. We extract the key representations $z^{mix}(i)$ from $x^{mix}(i)$ and feed this negative key into the queue (i.e., the memory bank) with the fixed size $|M|$.  Mixup samples serve as hard negative samples \cite{kalantidis2020hard, oh2023provable}, effectively improving the performance of instance discrimination loss. The proposed mixup-enhanced instance discrimination loss (MIDL) can be expressed by
\begin{align}
\mathcal{L}_{MIDL} = \frac{1}{|B^{low}|} \sum_{i \in B^{low}} \mathcal{L}_{MIDL}(i),
\end{align}
where
\begin{align}
    \mathcal{L}_{MIDL}(i) &= - \log \frac{\exp(z^1(i)\cdot z^2_+(i)/\tau_{m})}{\exp(z^1(i)\cdot z^2_+(i)/\tau_{m}) + \sum_{z^{mix}_{-} \in M}\exp(z^1(i)\cdot z^{mix}_{-}/\tau_{m})},
    \label{eq: instance discrimination loss}
\end{align}
where $M$ denotes the memory bank, $\tau_{m}$ denotes the temperature parameter, $z_+^2(i)$ denotes the representation of a positive key from a different view of the query representation $z^1(i)$ and $z^{mix}_{-}$ denotes the representation of a negative key sourced from the memory bank. This loss significantly enhances discrimination between low-confidence labeled samples in a self-supervised manner.

\subsection{Training with Semi-supervised Learning}
\label{sec: training with semi-supervised learning}

{\bf Balanced Classifier.}
Various methods \cite{zhou2020bbn,li2023sfa,wang2021ride} have implemented balanced classifiers to mitigate class bias towards head classes. Our method also employs a balanced classifier trained with Balanced Softmax \cite{ren2020balsoftmax}. We input the mixup image $x^{mix}(i)$ into the feature extractor $f$, followed by the balanced classifier, to produce the logit $\eta^{(mix, b)}(i)=\left[\eta_1^{(mix, b)}(i), ..., \eta^{(mix, b)}_{K}(i) \right]$, where $k$ is the class index. The loss function for Balanced Softmax is expressed as
\begin{align}
    \mathcal{L}_{BS}(i) &= - \log \frac{n_k \exp\left(\eta^{(mix, b)}_{k}(i)\right)}{\sum_{k=1}^{K} n_k \exp\left(\eta^{(mix, b)}_{k}(i)\right)},
    \label{eq: balanced softmax}
\end{align}
where $n_k$ denotes the number of samples of class $k$ among the data samples selected by the proposed DaCC. During the warmup phase, we determine $n_k$ based on the distribution of the training dataset $\mathcal{D}$. 

\noindent {\bf Total Loss.}
To effectively utilize noisy samples, we consider them as unlabeled data and train the model under a semi-supervised learning (SSL) framework \cite{berthelot2019mixmatch}. The total training loss is
\begin{align}
    \mathcal{L} &= \mathcal{L}_{MixMatch} + \mathcal{L}_{BMixMatch} + \lambda_{SBCL}\mathcal{L}_{SBCL} + \lambda_{MIDL}\mathcal{L}_{MIDL}, 
    \label{eq: total training loss}
\end{align}
where $\mathcal{L}_{MixMatch}$ denotes the MixMatch loss for the conventional classifier, $\mathcal{L}_{BMixMatch}$ denotes the MixMatch loss with $\mathcal{L}_{BS}$ for the balanced classifier,  and $\lambda_{SBCL}$ and $\lambda_{MIDL}$ are the coefficients for SBCL and MIDL, respectively. Algorithm \ref{alg: algorithm} presents the summary of our DaSC training framework. During the test time, we use the average of outputs from the classifiers $g^c$ and $g^b$ as the final model output. Furthermore, following previous studies \cite{li2020dividemix,wei2021robust,li2023sfa}, we employ a co-training framework against confirmation bias.

\begin{algorithm}[!]
	\caption{DaSC framework} 
    \textbf{Input}: Training dataset $\mathcal{D}=\{(x_i, \tilde{y}_i)\}^N_{i=1}$, model parameters $\theta$, warmup epochs $T_0$, warmup epochs for semi-supervised balanced contrastive loss $T_{SBCL}$
	\begin{algorithmic}[1]
		\For {$t=1,\ldots ,T_0$}
            \State $\mathcal{L}=\mathcal{L}_{CE}+\mathcal{L}_{BS}+\mathbb{I}(t>T_{SBCL})\lambda_{SBCL}\mathcal{L}_{SBCL}$
            \State Update $\theta$
        \EndFor
        \For {$t=T_0+1,\ldots ,T$}
            \State Compute the confidence threshold $\tau=\Phi^t\hat{\tau}$
            \State Compute the class centroids $c$ by Eqs. (1)-(3)
            \For {$i=1, \ldots, N$}
                \State Calculate the assignment probability $\Gamma$ by Eq. (4)
            \EndFor
            \State $\mathcal{D}^{clean}, \mathcal{D}^{noisy}$ = GMM$\left(\Gamma\right)$
            
            \State $\mathcal{L}=\mathcal{L}_{MixMatch}+\mathcal{L}_{BMixMatch}+\lambda_{SBCL}\mathcal{L}_{SBCL}+\lambda_{MIDL}\mathcal{L}_{MIDL}$
			\State Update $\theta$
		\EndFor
	\end{algorithmic} 
    \label{alg: algorithm}
\end{algorithm}

\section{Experiments}
\label{sec: experiments}

\subsection{Setup}
\label{sec: datasets and comparison methods}
{\bf Datasets.} The proposed method was evaluated on long-tailed version of the CIFAR-10 and CIFAR-100 datasets \cite{krizhevsky2009cifar}, with synthetic noise introduced to the labels. Additionally, we used real-world noise datasets including CIFAR-10N, CIFAR-100N \cite{wei2021learning}, and Red mini-ImageNet \cite{jiang2020beyond} for evaluation.

For the synthetic CIFAR-10 and CIFAR-100 datasets, we followed the experimental setup in \cite{lu2023tabasco}. We initially created long-tailed version of the CIFAR datasets by setting the number of samples for the $k$-th class as $N_k=\max_{k}|\mathcal{D}_k|\cdot\rho^{\frac{k-1}{K-1}}$, where $\rho$ is the imbalance ratio and
$K$ indicates the number of classes. Subsequently, we introduced two types of synthetic noise into the datasets: symmetric (Sym.) and asymmetric (Asym.) label noise. Symmetric noise involves randomly assigning labels to any other class based on a specified noise ratio. Asymmetric noise, on the other hand, changes the labels to semantically related classes (e.g., truck $\rightarrow$ automobile), taking class information into account. In our experiments, we used noise ratios of 0.4 and 0.6 for symmetric noise and 0.2 and 0.4 for asymmetric noise.

For CIFAR-10N (10N) and CIFAR-100N (100N) \cite{wei2021learning}, we modified the CIFAR datasets to the aforementioned long-tailed version and then introduced noisy labels created by human annotations. Red mini-ImageNet (Red) \cite{jiang2020beyond} is a smaller version of ImageNet \cite{russakovsky2015imagenet} that includes real-world noisy web labels with specific ratios. In this study, we used web labels with noise ratios of 0.2 and 0.4 for Red mini-ImageNet.

\noindent {\bf Baselines.} The following baseline methods were compared with our method. Methods addressing long-tailed distribution include LA \cite{menon2021logitadjust}, LDAM \cite{cao2019ldam}, IB Loss \cite{park2021influence}, and BBN \cite{zhou2020bbn}. Methods for dealing with noisy labels include DivideMix \cite{li2020dividemix}, UNICON \cite{karim2022unicon}, and TCL \cite{huang2023twin}. The latest methods addressing both noisy labels and long-tailed distribution include Meta-weight Net \cite{shu2019metaweight}, RoLT \cite{wei2021robust}, HAR \cite{cao2021har}, ULC \cite{huang2022uncertainty}, SFA \cite{li2023sfa}, and TABASCO \cite{lu2023tabasco}.

\subsection{Implementation Details}
\label{sec: implementation details}
We used PreAct ResNet18 and ResNet18 \cite{he2009learning} as backbone networks for the experiments on the CIFAR and Red mini-ImageNet datasets, respectively. The model was trained for 100 epochs using a stochastic gradient descent optimizer with an initial learning rate of 0.02, a momentum of 0.9, and a batch size of 64. Weight decay of $1 \times 10^{-4}$ and $5 \times 10^{-5}$ were applied for the CIFAR and Red mini-ImageNet datasets, respectively. Contrastive loss was introduced after the first 10 epochs, and correctly labeled samples were detected after the first 30 epochs. Data augmentation techniques included random horizontal flips, random cropping, and RandAugment \cite{cubuk2020rand}. Table \ref{tab: hyperparameters} provides the configuration parameters of our DaSC.

\begin{table}[tbh]
    \centering
    \caption{The configuration parameters used for DaSC.}
    \begin{adjustbox}{width=0.65\textwidth, center}
    \begin{tabular}{l|c}
        \hline
        \centering
        Hyperparameter & Value \\
        \hline
        The coefficient $\lambda_{SBCL}$ for SBCL & 0.5 \\
        The temperature $\tau_{s}$ for SBCL & 0.1 \\
        The coefficient $\lambda_{MIDL}$ for MIDL & 0.3 \\
        The temperature $\tau_{m}$ for MIDL & 0.5 \\
        The temperature parameter $\tau_{T}$ & 0.1 \\
        The confidence threshold $\tau_c$ for sample grouping & 0.9 \\
        \hline
    \end{tabular}
    \end{adjustbox}
    \label{tab: hyperparameters}
\end{table}

\begin{table}[t]
    \centering
    \caption{Performance of the proposed DaSC on the long-tailed version of CIFAR datasets with (a) a symmetric noise and (b) an asymmetric noise.}
    \subfloat[Symmetric noise]{
    \begin{adjustbox}{width=0.475\linewidth}
    \begin{tabular}{l|lcccc}
        \Xhline{1.5pt}
        \multicolumn{2}{l}{Dataset} &  \multicolumn{2}{c}{CIFAR-10}& \multicolumn{2}{c}{CIFAR-100}\\
        \hline
        \multicolumn{2}{l}{Imbalance Ratio} & \multicolumn{4}{c}{0.1} \\
        \hline
        \multicolumn{2}{l}{Noise Ratio} & 0.4 & 0.6 & 0.4 & 0.6 \\
        \hline
        Baseline & CE & 71.67 & 61.16 & 34.53 & 23.63 \\
        \hline
        \multirow{4}{*}{LT} & LA \cite{menon2021logitadjust} & 70.56 & 54.92 & 29.07 & 23.21 \\
        & LDAM \cite{cao2019ldam} & 70.53 & 61.97 & 31.30 & 23.13 \\
        & IB \cite{park2021influence} & 73.24 & 62.62 & 32.40 & 25.84 \\
        & BBN \cite{zhou2020bbn} & 71.16 & 56.38 & 30.48 & 23.00 \\
        \hline
        \multirow{3}{*}{NL} & DivideMix \cite{li2020dividemix} & 82.67 & 80.17 & 54.71 & 44.98 \\
        & UNICON \cite{karim2022unicon} & 84.25 & 82.29 & 52.34 & 45.87 \\
        & TCL \cite{huang2023twin} & 79.55 & 72.13 & 51.49 & 41.19 \\
        \hline
        \multirow{7}{*}{NL-LT} & MW-Net \cite{shu2019metaweight} & 70.90 & 59.85 & 32.03 & 21.71\\
        & RoLT \cite{wei2021robust} & 81.62 & 76.58 & 42.95 & 32.59 \\
        & HAR \cite{cao2021har} & 77.44 & 63.75 & 38.17 & 26.09 \\
        & ULC \cite{huang2022uncertainty} & 84.46 & 83.25 & 54.91 & 44.66 \\
        & SFA \cite{li2023sfa} & 86.81 & 82.89 & 56.70 & 47.71 \\
        & TABASCO \cite{lu2023tabasco} & 85.53 & 84.83 & 56.52 & 45.98 \\
        \cline{2-6}
        & DaSC & \bf{89.04} & \bf{87.12} & \bf{61.85} & \bf{54.40} \\
        \Xhline{1.5pt}
    \end{tabular}
    \label{tab: symmetric on cifar} 
    \end{adjustbox}}
    \quad
    \subfloat[Asymmetric noise]{
    \begin{adjustbox}{width=0.475\linewidth}
    \begin{tabular}{l|lcccc}
        \Xhline{1.5pt}
        \multicolumn{2}{l}{Dataset} &  \multicolumn{2}{c}{CIFAR-10}& \multicolumn{2}{c}{CIFAR-100}\\
        \hline
        \multicolumn{2}{l}{Imbalance Ratio} & \multicolumn{4}{c}{0.1} \\
        \hline
        \multicolumn{2}{l}{Noise Ratio} & 0.2 & 0.4 & 0.2 & 0.4 \\
        \hline
        Baseline & CE & 79.90 &  62.88 & 44.45 & 32.05 \\
        \hline
        \multirow{4}{*}{LT} & LA \cite{menon2021logitadjust} & 71.49 & 59.88 & 39.34 & 28.49 \\
        & LDAM \cite{cao2019ldam} & 74.58 & 62.29 & 40.06 & 33.26 \\
        & IB \cite{park2021influence} & 73.49 & 58.36 & 45.02 & 35.25 \\
        & BBN \cite{zhou2020bbn} & 75.57 & 60.91 & 39.16 & 31.03 \\
        \hline
        \multirow{3}{*}{NL} & DivideMix \cite{li2020dividemix} & 80.92 & 69.35 & 58.09 & 41.99 \\
        & UNICON \cite{karim2022unicon} & 72.81 & 69.04 & 55.99 & 44.70 \\
        & TCL \cite{huang2023twin} & 84.55 & 72.97 & 58.61 & 44.75\\
        \hline
        \multirow{7}{*}{NL-LT} & MW-Net \cite{shu2019metaweight} & 79.34 & 65.49 & 42.52 & 30.42 \\
        & RoLT \cite{wei2021robust} & 73.30 & 58.29 & 48.19 & 39.32 \\
        & HAR \cite{cao2021har} & 82.85 & 69.19 & 48.50 & 33.20 \\
        & ULC \cite{huang2022uncertainty} & 74.07 & 73.19 & 54.45 & 43.20 \\
        & SFA \cite{li2023sfa} & 87.70 & 78.13 & 57.16 & 46.09 \\ 
        & TABASCO \cite{lu2023tabasco} & 82.10 & 80.57 & 59.39 & 50.51 \\
        \cline{2-6}
        & DaSC & \bf{89.89} & \bf{88.85} & \bf{63.22} & \bf{53.66} \\
        \Xhline{1.5pt}
    \end{tabular}
    \label{tab: asymmetric on cifar} 
    \end{adjustbox}}
    \label{tab: cifar}
\end{table}

\begin{table}[t]
    \centering
    \captionof{table}{Performance of the proposed DaSC on the long-tailed version of CIFAR datasets and Red mini-ImageNet dataset.}
    \begin{adjustbox}{width=0.5\columnwidth}
    \begin{tabular}{l|lcccc}
        \Xhline{1.5pt}
        \multicolumn{2}{l}{Dataset} &  \multicolumn{2}{c}{Red}& 10N & 100N\\
        \hline
        \multicolumn{2}{l}{Imbalance Ratio} & \multicolumn{2}{c}{$\approx$ 0.1} & \multicolumn{2}{c}{0.1} \\
        \hline
        \multicolumn{2}{l}{Noise Ratio} & 0.2 & 0.4 & \multicolumn{2}{c}{$\approx$ 0.4} \\
        \hline
        Baseline & CE & 40.42 & 31.46 & 60.44 & 38.10 \\
        \hline
        \multirow{4}{*}{LT} & LA \cite{menon2021logitadjust} & 26.82 & 25.88 & 65.74 & 36.50 \\
        & LDAM \cite{cao2019ldam} & 26.64 & 23.46 & 62.50 & 38.48 \\
        & IB \cite{park2021influence} & 23.80 & 22.08 & 65.91 & 42.48 \\
        & BBN \cite{zhou2020bbn} & 42.18 & 42.64 & 68.42 & 37.22 \\
        \hline
        \multirow{3}{*}{NL} & DivideMix \cite{li2020dividemix} & 48.76 & 48.96 & 67.85 & 44.25 \\
        & UNICON \cite{karim2022unicon} & 40.18 & 41.64 & 69.54 & 51.93 \\
        & TCL \cite{huang2023twin} & 37.12 & 37.76 & 73.46 & 51.74 \\
        \hline
        \multirow{7}{*}{NL-LT} & MW-Net \cite{shu2019metaweight} & 42.66 & 40.26 & 69.73 & 44.20 \\
        & RoLT \cite{wei2021robust} & 22.56 & 24.22 & 75.24 & 46.61 \\
        & HAR \cite{cao2021har} & 46.61 & 38.71 & 74.97 & 44.54 \\
        & ULC \cite{huang2022uncertainty} & 48.12 & 47.06 & 75.71 & 51.72 \\
        & SFA \cite{li2023sfa} & 42.50 & 43.62 & 84.57 & 53.68 \\
        & TABASCO \cite{lu2023tabasco} & 50.20 & 49.68 & 80.61 & 53.83 \\ 
        \cline{2-6}
        & DaSC & \bf{52.24} & \bf{50.58} & \bf{86.01} & \bf{58.10} \\
        \Xhline{1.5pt}
    \end{tabular}
    \end{adjustbox}
    \label{tab: real-world noise dataset}
\end{table}

\subsection{Main Results}
\label{sec: experiment results}

{\bf Results on CIFAR Datasets.}
Tables \ref{tab: symmetric on cifar} and \ref{tab: asymmetric on cifar} display the performance of the proposed DaSC on long-tailed noisy versions of the CIFAR datasets. We examine both symmetric and asymmetric noise in Tables \ref{tab: symmetric on cifar} and \ref{tab: asymmetric on cifar}, respectively. The proposed DaSC consistently outperforms other baseline methods in all experimental setups. For CIFAR-10 with a symmetric noise ratio of 0.4, DaSC achieves performance gains of 2.23\% and 3.51\% over the current best methods, SFA and TABASCO, respectively. These performance gains tend to increase as the noise ratio and the number of classes rise. On CIFAR-100 with a symmetric noise ratio of 0.6, DaSC achieves substantial performance gains of 6.69\% and 8.42\% over SFA and TABASCO, respectively.

\noindent {\bf Results on Real-World Datasets.}
Table \ref{tab: real-world noise dataset} shows the performance of DaSC on real-world datasets with long-tailed distributions. These datasets are more challenging than those with synthetic labeling noise because samples are more prone to being mislabeled into semantically similar classes, and noisy labels are unevenly distributed. Notably, our DaSC method achieves significant performance gains on real-world datasets created by human annotations, aligning with the experimental results observed in the previous experiments. Our DaSC achieves a performance gain of 2.04\% over the existing best method, TABASCO \cite{lu2023tabasco} with a noise ratio of 0.2 on the Red mini-ImageNet dataset. Additionally, DaSC outperforms TABASCO by 5.40\% on CIFAR-10N and by 4.27\% on CIFAR-100N.

\begin{figure}[t]
    \centering
    \begin{minipage}[b]{0.49\textwidth}
        \captionof{table}{Ablation study on the long-tailed version of CIFAR.}
        \begin{adjustbox}{width=1\columnwidth}
        \begin{tabular}{c|c|c|c|cccc}
            \Xhline{1.5pt}
            \multicolumn{4}{c|}{Main Component} &  \multicolumn{2}{c}{CIFAR-10}& \multicolumn{2}{c}{CIFAR-100}\\
            \hline
            \multicolumn{2}{c|}{Sample Selection} & \multicolumn{2}{c|}{Contrastive Learning} & Sym. & Asym. & Sym. & Asym.\\
            \hline
            DaCC & TS & SBCL & MIDL & 0.4 & 0.2 & 0.4 & 0.2 \\
            \hline
             &  & & & 85.28 & 84.76 & 54.76 & 58.35 \\
            \checkmark &  & & & 87.13 & 87.98 & 57.81 & 59.93 \\
            \checkmark & \checkmark & & & 87.53 & 88.48 & 58.23 & 61.28 \\
            \checkmark & \checkmark & \checkmark & & 88.29 & 89.14 & 60.99 & 61.94 \\
            \checkmark & \checkmark & \checkmark & \checkmark &  \bf{89.04} & \bf{89.89} &\bf{61.85} & \bf{63.22} \\
            \Xhline{1.5pt}
        \end{tabular}
        \end{adjustbox}
        \label{tab: ablation study}
    \end{minipage}
    \begin{minipage}[b]{0.475\textwidth}
        \captionof{table}{Performance comparison with different types of negative keys.}
        \vspace{0.1cm}
        \begin{adjustbox}{width=1\columnwidth, center}
        \begin{tabular}{c|cccc}
            \Xhline{1.5pt}
            \multirow{3}{*}{Negative Key} &  \multicolumn{2}{c}{CIFAR-10}& \multicolumn{2}{c}{CIFAR-100}\\
            \cline{2-5}
             & Sym. & Asym. & Sym. & Asym.\\
             & 0.4 & 0.2 & 0.4 & 0.2 \\
            \hline
             Without Mixup & 88.25 & 89.35 & 61.22 & 62.50 \\
             With Mixup & \bf{89.04} & \bf{89.89} &\bf{61.85} & \bf{63.22} \\
            \Xhline{1.5pt}
        \end{tabular}
        \end{adjustbox}
        \label{tab: negative key}
    \end{minipage}
\end{figure}

\subsection{Performance Analysis}
\label{sec: ablation study and analysis}
{\bf Ablation Study.} We conducted an ablation study to analyze the impact of the main components in the proposed DaSC. Table \ref{tab: ablation study} presents the experimental results on CIFAR datasets with various noise types and ratios. DaCC offers significant performance gains over the baseline, confirming the effectiveness of the proposed centroid estimation method. Specifically, it achieves a 1.85\% performance gain on CIFAR-10 with a symmetric noise ratio of 0.4. Incorporating TS into DaCC leads to further performance improvement of 0.4\% gain under the same setup. Adding SBCL further enhances performance by 0.76\%. The combination of SBCL and MIDL provides a 1.51\% performance gain, resulting in a total performance gain of 3.76\% over the baseline. These performance gains are consistent across various datasets and noise ratios, demonstrating the robustness and versatility of the proposed DaSC.

\noindent {\bf Hyperparameters.} 
\label{sec: hyperparameters} 
The proposed DaSC performs consistently well across various hyperparameter configurations. The experimental results with different hyperparameter configurations are provided in the supplementary material.

\noindent {\bf Generation of Negative Keys for MIDL.} Table \ref{tab: negative key} illustrates the effect of using mixup samples to generate negative keys for MIDL. In the absence of mixup samples, low-confidence samples are used as negative keys. Notably, using mixup samples results in better performance compared to not using them, demonstrating that mixup samples provide more effective negative keys for contrastive learning.

\begin{figure}[t]
    \centering
    \begin{minipage}[b]{0.49\textwidth}
        \begin{subfigure}[t]{0.49\columnwidth}
            \centering
            \includegraphics[width=1\columnwidth]{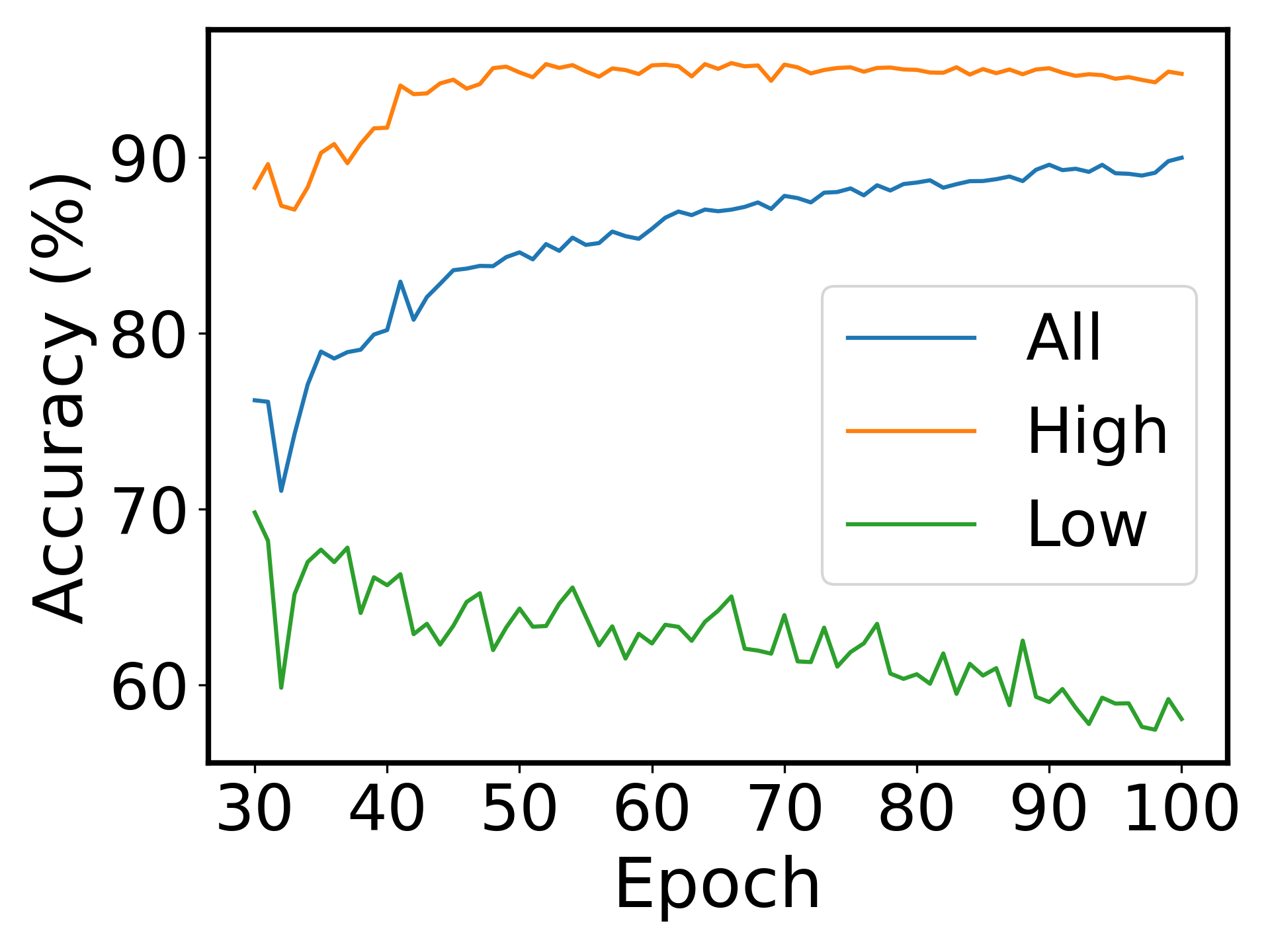}
            \caption[]{Sym. 0.6}
        \end{subfigure}
        \begin{subfigure}[t]{0.49\columnwidth}
            \centering
            \includegraphics[width=1\columnwidth]{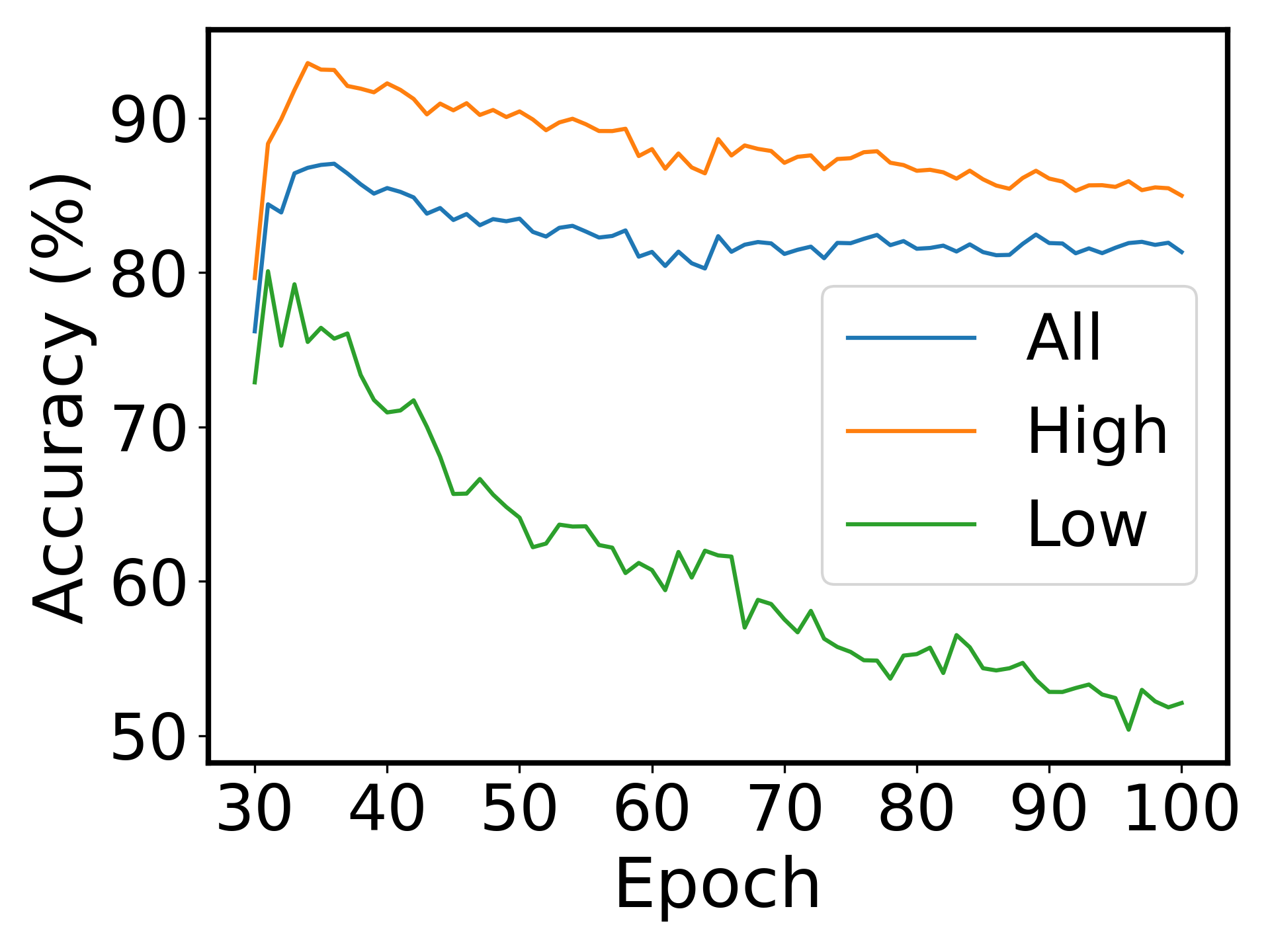}
            \caption[]{Asym. 0.4}
        \end{subfigure}
        \caption{Accuracy of pseudo-labels for different sample groups.}
        \label{fig: pseudo label accuracy}
    \end{minipage}
    \begin{minipage}[b]{0.475\textwidth}
    \captionof{table}{Performance of DaSC when different sample groups are used for SBCL and MIDL.}
    \centering
    \begin{adjustbox}{width=1\columnwidth}
    \begin{tabular}{c|c|cccc}
        \Xhline{1.5pt}
        \multicolumn{2}{c|}{Sample Set} &  \multicolumn{2}{c}{CIFAR-10}& \multicolumn{2}{c}{CIFAR-100}\\
        \hline
        \multirow{2}{*}{SBCL} & \multirow{2}{*}{MIDL} & Sym. & Asym. & Sym. & Asym.\\
        \cline{3-6}
        & & 0.4 & 0.2 & 0.4 & 0.2 \\
        \hline
        All & All & 88.19 & 89.20 & 60.86 & 62.20\\
        High & All & 88.45 & 89.52 & 61.30 & 62.69\\
        All & Low & 88.76 & 89.46 & 61.25 & 62.77 \\
        High & Low &  \bf{89.04} & \bf{89.89} &\bf{61.85} & \bf{63.22} \\
        \Xhline{1.5pt}
    \end{tabular}
    \end{adjustbox}
    \label{tab: sample set}
    \end{minipage}
\end{figure}

\noindent {\bf Analysis of Confidence-Aware Contrastive Learning.} First, Fig. \ref{fig: pseudo label accuracy} shows the accuracy of pseudo-labels relative to true labels for different sample groups. The terms {\it `All'}, {\it `High'}, and {\it `Low'} refer to using all samples, a high-confidence sample group, and a low-confidence sample group, respectively. As expected, the high-confidence samples align more closely with the true labels, while low-confidence samples tend to be less consistent with the true labels.

Next, Table \ref{tab: sample set} evaluates the effect of using different sample groups for SBCL and MIDL. Our results confirm that using the high-confidence sample group for SBCL and the low-confidence sample group for MIDL yields the best performance among the configurations tested in the experiments. This highlights the effectiveness of our contrastive learning strategy, which efficiently utilizes each sample group for representation learning.

\begin{figure}[t]
    \centering
    \begin{minipage}{.49\textwidth}
        \begin{subfigure}[b]{0.49\columnwidth}
            \centering
            \includegraphics[width=1\columnwidth]{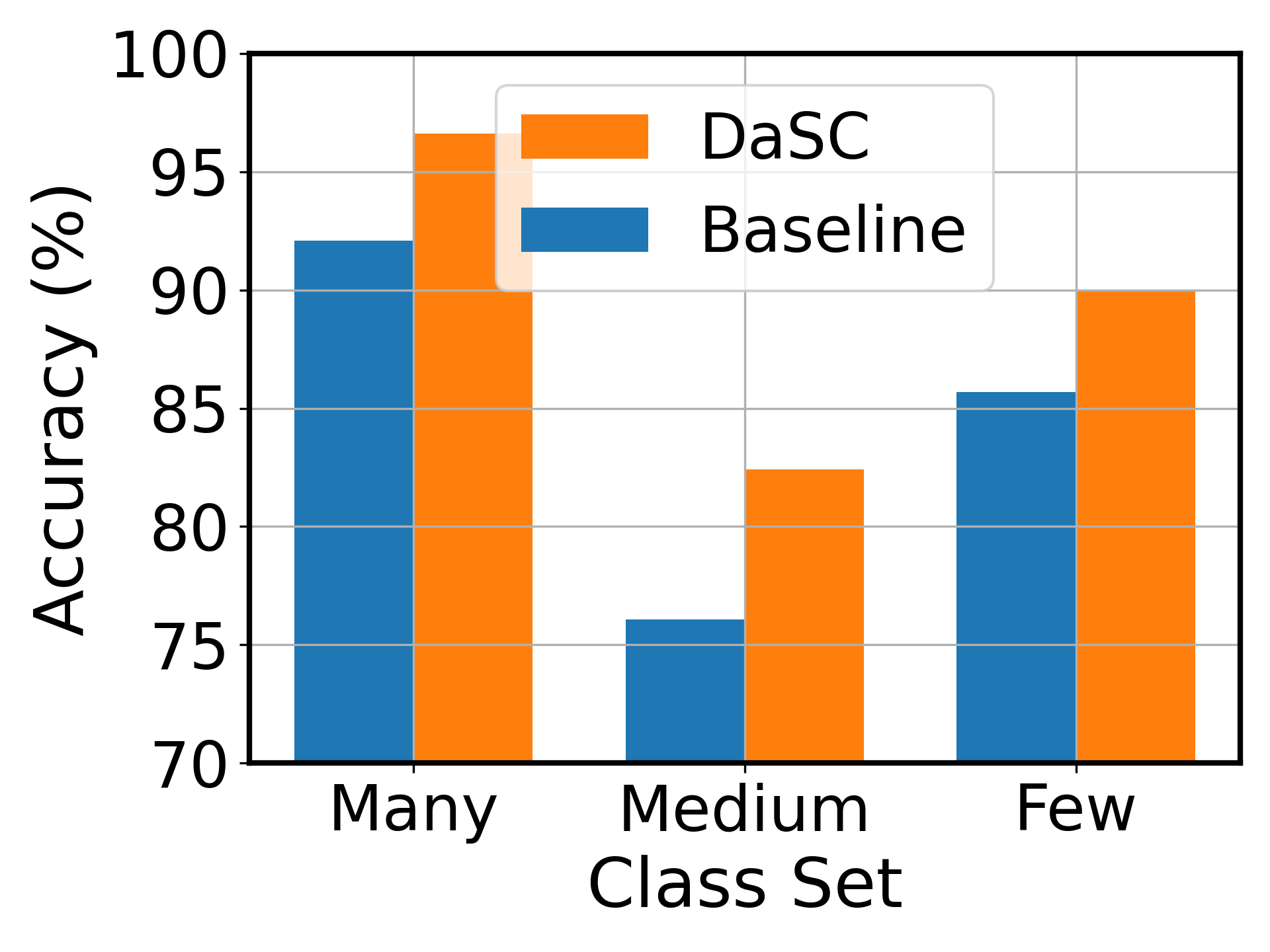}
            \caption[]{Sym. 0.6}
        \end{subfigure}
        \begin{subfigure}[b]{0.49\columnwidth}
            \centering
            \includegraphics[width=1\columnwidth]{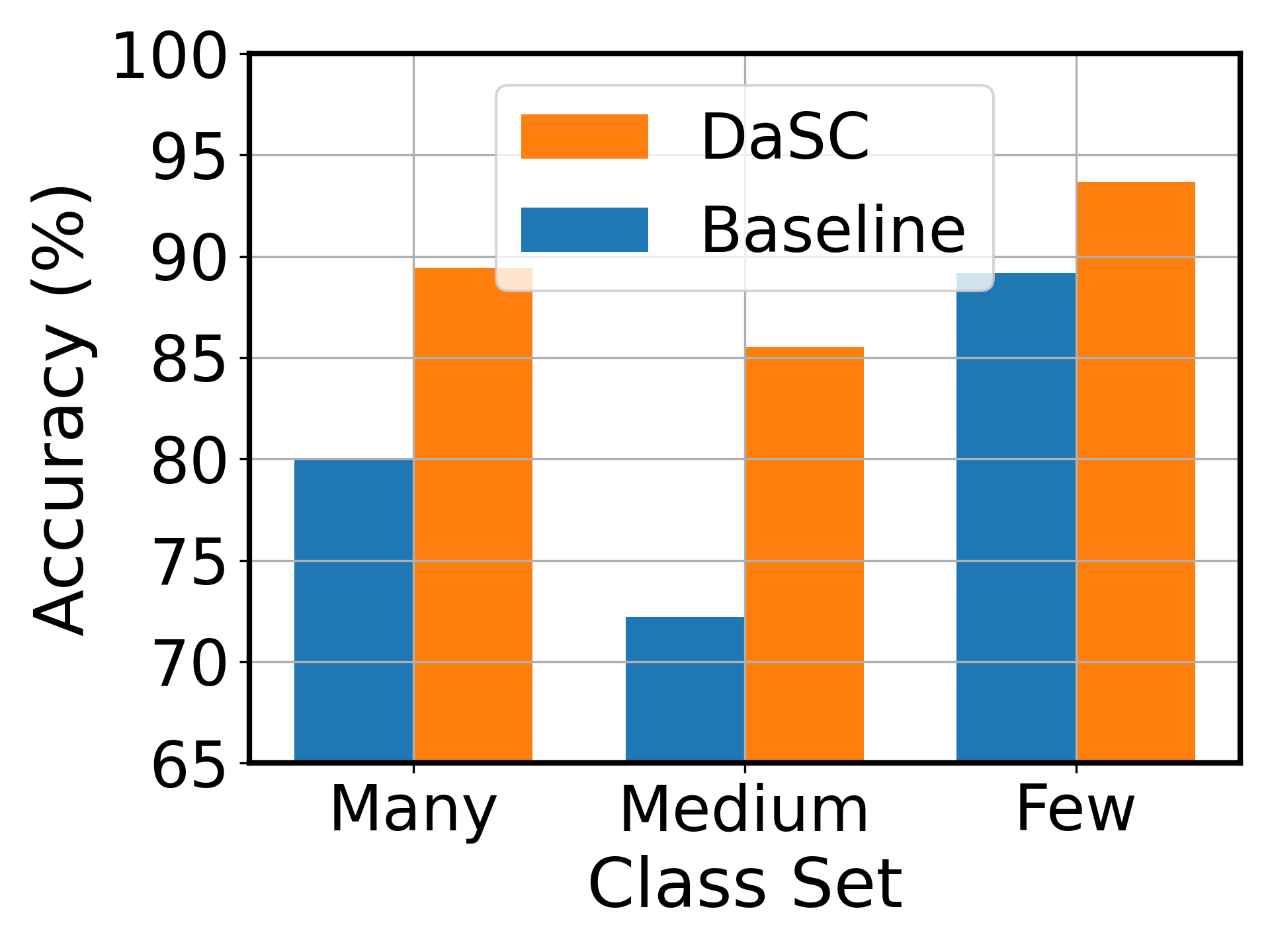}
            \caption[]{Asym. 0.4}
        \end{subfigure}
        \caption[]{Performance of the DaSC compared with the baseline method for different class sets including `Many', `Medium', and `Few'. }
        \label{fig: performance per class sets}
    \end{minipage}
    \begin{minipage}{.49\textwidth}
        \begin{subfigure}[b]{0.485\columnwidth}
            \centering
            \includegraphics[width=1\columnwidth]{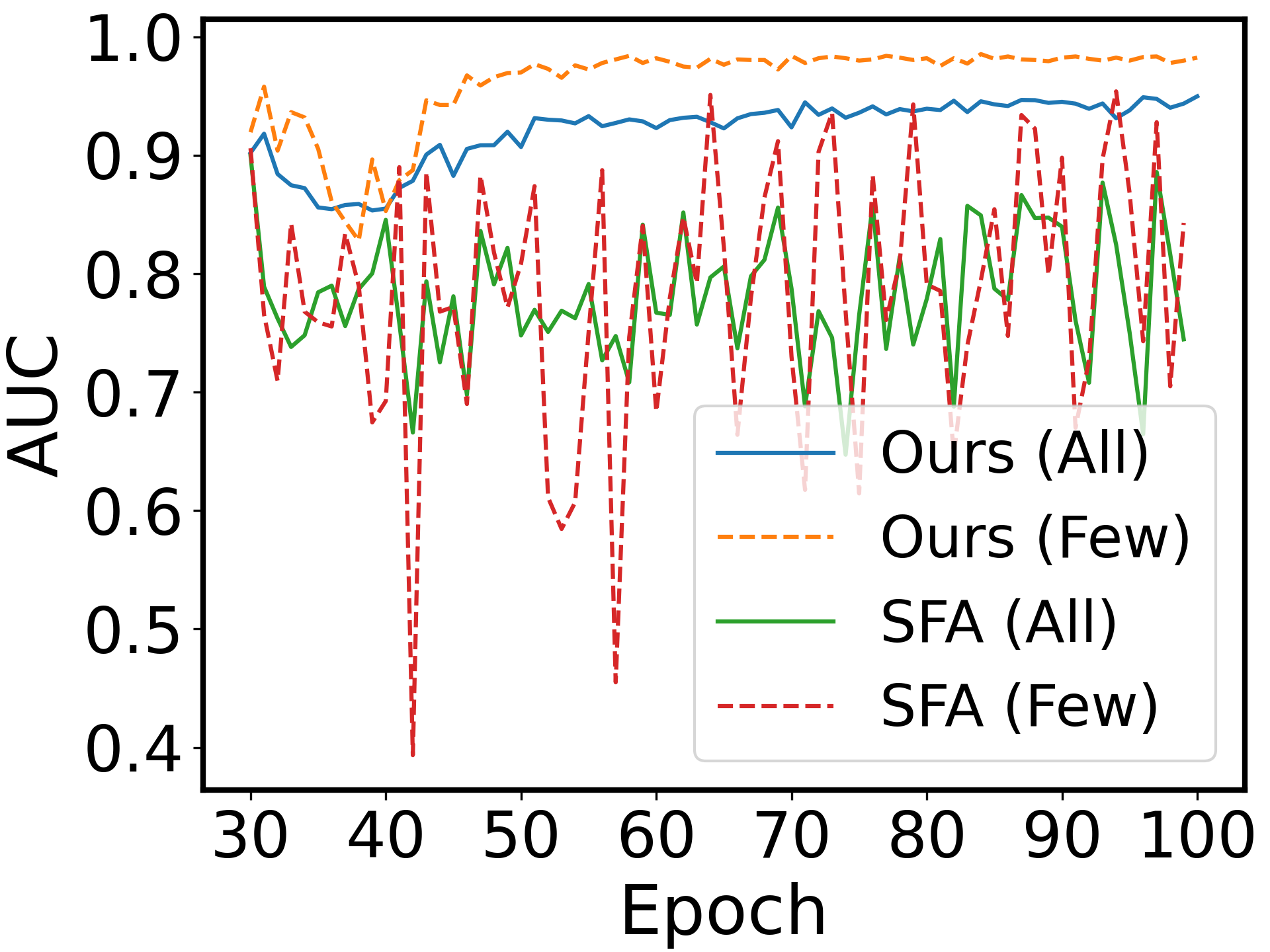}
            \caption[]{Sym. 0.6}
        \end{subfigure}
        \begin{subfigure}[b]{0.485\columnwidth}
            \centering
            \includegraphics[width=1\columnwidth]{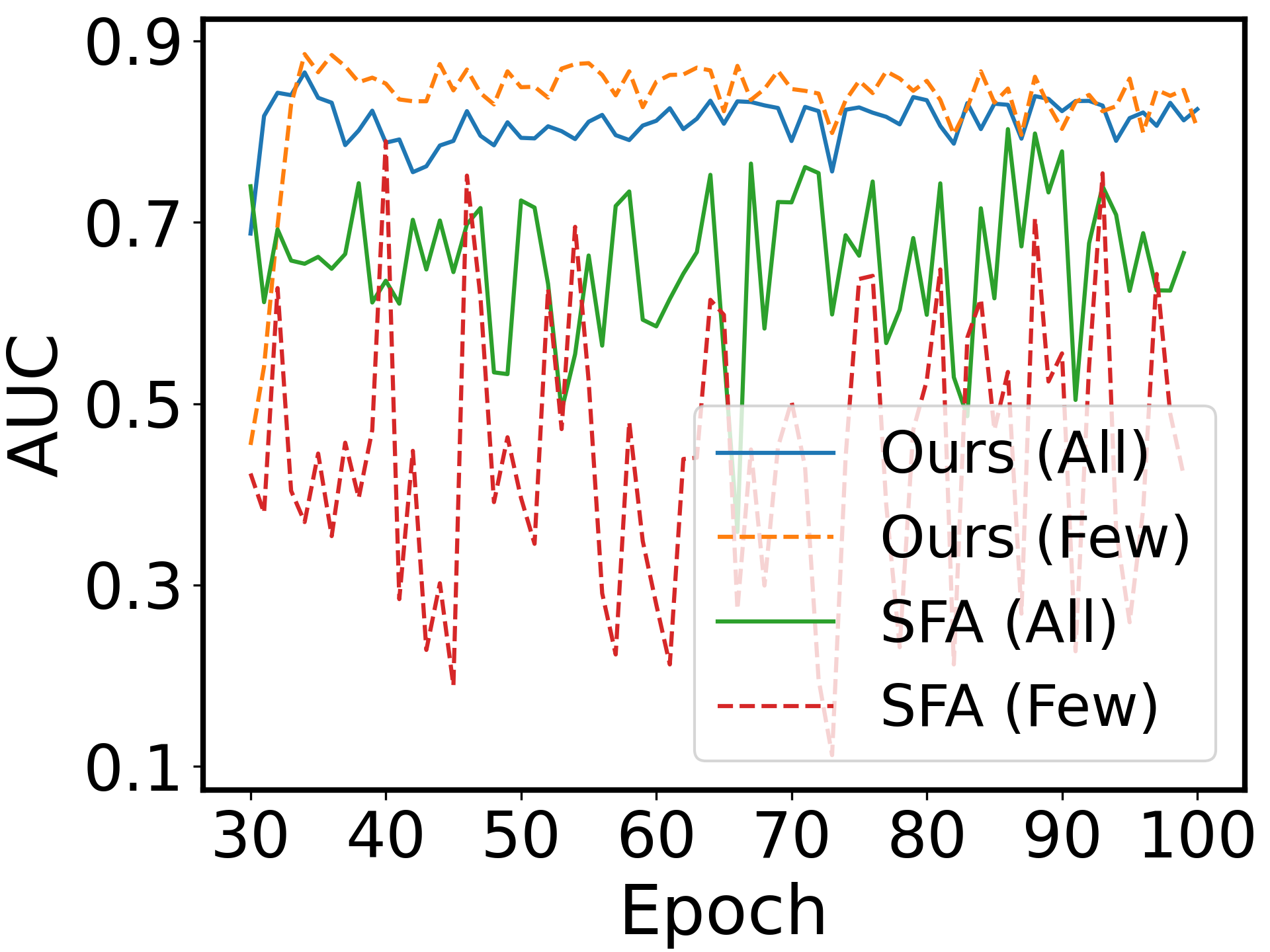}
            \caption[]{Asym. 0.4}
        \end{subfigure}
        \caption[]{AUC of the proposed DaSC in comparison of SFA for noisy label detection. We present the performance for `All' and `Few' class sets.}
    \label{fig: AUC}
    \end{minipage}
\end{figure}

\noindent {\bf Performance Evaluation on Different Class Sets.} 
In Fig. \ref{fig: performance per class sets}, we compare the performance of DaSC with that of the baseline across different class sets: {\it `Many'}, {\it `Medium'}, and {\it `Few'}. This categorization is based on the framework presented in previous work \cite{liu2019large}. We evaluate both methods on the long-tailed version of CIFAR-10 with symmetric noise of 0.6 and asymmetric noise of 0.4. The proposed DaSC method significantly improves performance across all class sets. We observe that DaSC enhances performance not only for the Few and Medium class sets but also for the Many class set. These improvements suggest that DaSC effectively enhances data representation using confidence-aware contrastive learning, leading to better recognition of samples across the Many, Medium, and Few class sets.

\noindent {\bf Accuracy of Detecting Noisy Labels.}
Figure \ref{fig: AUC} illustrates the area under the curve (AUC) for DaSC in detecting noisy labels using centroids for each epoch. We compare our method with the SFA, with both methods trained on the long-tailed version of CIFAR-10, incorporating symmetric label noise of 0.6 and asymmetric label noise of 0.4. Additionally, we present the AUC results for both methods specifically for the `Few' class set. Experimental results demonstrate that our method consistently outperforms SFA throughout the entire training process. Notably, while the accuracy of SFA fluctuates across epochs, our method shows a smooth improvement. In the Few class set, SFA struggles to converge until the end of the training, whereas our method maintains consistently strong detection performance. These results demonstrate the effectiveness of our method in detecting correctly labeled samples, even within the Few class set.

\section{Conclusion}
\label{sec: conclusion}
In this study, we proposed a novel robust learning framework, DaSC, to address the challenges of long-tailed class distribution and noisy label simultaneously. Recent approaches have employed feature-based noisy sample selection methods that rely on class centroids computed using high-confidence samples within each target class. In contrast, we introduced DaCC, which estimates class centroids using samples from all other classes. To obtain high-quality class centroids, we adaptively weight the samples based on a confidence score refined by temperature scaling. Additionally, we introduced a confidence-aware contrastive learning framework to enhance representation quality. We first categorized each training sample into a high-confidence sample and a low-confidence sample. Then, we computed SBCL for high-confidence samples and MIDL for low-confidence samples. SBCL incorporates class-balanced contrastive learning loss using class information from high-confidence samples, while MIDL employs mixup samples derived from low-confidence samples as negative keys for contrastive learning. Our extensive experiments demonstrated that the proposed DaSC achieves state-of-the-art performance in widely used benchmarks. 

\bibliographystyle{splncs04}
\bibliography{egbib}
%
%
%

\clearpage
\appendix

\section{Experimental Results with Higher Imbalance Ratio}
In this section, we provide the performance of the proposed method compared to other baselines on long-tailed CIFAR and mini-ImageNet datasets with an imbalance ratio of 0.01 in Tables \ref{tab: imb 100 cifar} and \ref{tab: imb 100 real-world noise dataset}. The performances of other methods were taken from \cite{lu2023tabasco}. For methods not reported in \cite{lu2023tabasco}, we reproduced their performances. For all experimental setups, the proposed DaSC consistently outperforms other baseline methods. Specifically, on CIFAR-10 with symmetric noise of 0.4, DaSC achieves 2.37$\%$ and 8.01$\%$ better performance over SFA and TABASCO, respectively. Similarly, for CIFAR-10 with asymmetric noise of 0.2, DaSC outperforms SFA and TABASCO by 4.93$\%$ and 10.58$\%$, respectively. For CIFAR-10N with human annotations, DaSC achieves significant performance improvements of 6.97$\%$ and 6.07$\%$ over SFA and TABASCO, respectively. These results demonstrate the effectiveness and robustness of DaSC against extremely noisy labels and long-tailed distributions.

\begin{table}[h!]
    \centering
    \caption{Performance of the proposed method compared to baseline methods on the long-tailed version of CIFAR datasets with (a) symmetric noise and (b) asymmetric noise. The best results are shown in bold.}
    \subfloat[Symmetric noise]{
    \begin{adjustbox}{width=0.47\linewidth}
    \begin{tabular}{l|lcccc}
        \Xhline{1.5pt}
        \multicolumn{2}{l}{Dataset} &  \multicolumn{2}{c}{CIFAR-10}& \multicolumn{2}{c|}{CIFAR-100}\\
        \hline
        \multicolumn{2}{l}{Imbalance Ratio} & \multicolumn{4}{c}{0.01} \\
        \hline
        \multicolumn{2}{l}{Noise Ratio} & 0.4 & 0.6 & 0.4 & 0.6 \\
        \hline
        Baseline & CE & 47.81 & 28.04 & 21.99  & 15.51  \\
        \hline
        \multirow{4}{*}{LT} & LA \cite{menon2021logitadjust} & 42.63 & 36.37 & 21.54  & 13.14 \\
        & LDAM \cite{cao2019ldam} & 45.52 & 35.29 & 18.81 & 12.65  \\
        & IB \cite{park2021influence} & 49.07 & 32.54  & 20.34 & 12.10 \\
        & BBN \cite{zhou2020bbn} & 45.22 & 31.63 & 17.31 & 12.83 \\
        \hline
        \multirow{3}{*}{NL} & DivideMix \cite{li2020dividemix} & 32.42 & 34.73 & 36.20  & 26.29 \\
        & UNICON \cite{karim2022unicon} & 61.23 & 54.69  & 32.09  & 24.82 \\
        & TCL \cite{huang2023twin} & 56.13 & 45.88 & 33.29 & 24.39 \\
        \hline
        \multirow{7}{*}{NL-LT} & MW-Net \cite{shu2019metaweight} & 46.62 & 39.33 & 19.65 &  13.72 \\
        & RoLT \cite{wei2021robust} & 60.11 & 44.23 & 23.51 & 16.61 \\
        & HAR \cite{cao2021har} & 51.54 & 38.28 & 20.21 & 14.89 \\
        & ULC \cite{huang2022uncertainty} & 45.22 & 50.56 & 33.41 & 25.69 \\
        & SFA \cite{li2023sfa} & 67.98 & 54.70 & 37.69 & 30.02 \\
        & TABASCO \cite{lu2023tabasco} & 62.34 & 55.76 & 36.91 & 26.25 \\
        \cline{2-6}
        & DaSC & \bf{70.35} & \bf{58.49} &  \bf{41.12} & \bf{33.65} \\
        \Xhline{1.5pt}
    \end{tabular}
    \end{adjustbox}}
    \quad
    \subfloat[Asymmetric noise]{
    \begin{adjustbox}{width=0.48\linewidth}
    \begin{tabular}{l|lcccc}
        \Xhline{1.5pt}
        \multicolumn{2}{l}{Dataset} &  \multicolumn{2}{c}{CIFAR-10}& \multicolumn{2}{c|}{CIFAR-100}\\
        \hline
        \multicolumn{2}{l}{Imbalance Ratio} & \multicolumn{4}{c}{0.01} \\
        \hline
        \multicolumn{2}{l}{Noise Ratio} & 0.2 & 0.4 & 0.2 & 0.4 \\
        \hline
        Baseline & CE & 56.56 & 44.64 & 25.35 & 17.89 \\
        \hline
        \multirow{4}{*}{LT} & LA \cite{menon2021logitadjust} & 58.78 & 43.37 & 32.16 & 22.67 \\
        & LDAM \cite{cao2019ldam} & 61.25 & 40.85 & 29.22 & 18.65 \\
        & IB \cite{park2021influence} & 56.28 & 42.96 & 31.15 & 23.40 \\
        & BBN \cite{zhou2020bbn} & 54.51 & 51.15 & 25.19 & 21.68 \\
        \hline
        \multirow{3}{*}{NL} & DivideMix \cite{li2020dividemix} & 41.12 & 42.79 & 38.46 & 29.69 \\
        & UNICON \cite{karim2022unicon} & 53.53 & 34.05 & 34.14 & 30.72 \\
        & TCL \cite{huang2023twin} & 60.58 & 49.66 & 40.18 & 30.54 \\
        \hline
        \multirow{7}{*}{NL-LT} & MW-Net \cite{shu2019metaweight} & 62.19 & 45.21 & 27.56 & 20.40 \\
        & RoLT \cite{wei2021robust} & 54.81 & 50.26 & 32.96 & - \\
        & HAR \cite{cao2021har} & 62.42 & 51.97 & 27.90 & 20.03 \\
        & ULC \cite{huang2022uncertainty} & 41.14 & 22.73 & 34.07 & 25.04 \\
        & SFA \cite{li2023sfa} & 68.63 & 52.16 & 41.89 & 33.33 \\
        & TABASCO \cite{lu2023tabasco} & 62.98 & 54.04 & 40.35 & 33.15 \\
        \cline{2-6}
        & DaSC & \bf{73.56} & \bf{58.45} & \bf{43.52} & \bf{35.12} \\
        \Xhline{1.5pt}
    \end{tabular}
    \label{tab: imb 100 asymmetric on cifar} 
    \end{adjustbox}}
    \label{tab: imb 100 cifar}
\end{table}

\begin{table}[h!]
    \centering
    \caption{Performance of the proposed method compared to baseline methods on the long-tailed version of CIFAR with human annotations and Red mini-ImageNet dataset. The best results are shown in bold.}
    \begin{adjustbox}{width=0.5\columnwidth,center}
    \begin{tabular}{l|lcccc}
        \Xhline{1.5pt}
        \multicolumn{2}{l}{Dataset} &  \multicolumn{2}{c}{Red}& 10N & 100N\\
        \hline
        \multicolumn{2}{l}{Imbalance Ratio} & \multicolumn{2}{c}{$\approx$ 0.01} & \multicolumn{2}{c}{0.01} \\
        \hline
        \multicolumn{2}{l}{Noise Ratio} & 0.2 & 0.4 & \multicolumn{2}{c}{$\approx$ 0.4} \\
        \hline
        Baseline & CE & 30.88 & 31.46 & 49.31 & 25.28 \\
        \hline
        \multirow{4}{*}{LT} & LA \cite{menon2021logitadjust}& 10.32 & 9.56 & 50.09 & 26.39 \\
        & LDAM \cite{cao2019ldam} & 14.30 & 15.64 & 50.36 & 30.17 \\
        & IB \cite{park2021influence} & 16.72 & 16.34 & 56.41 & 31.55 \\
        & BBN \cite{zhou2020bbn} & 30.92 & 30.30 & 52.98 & 25.06 \\
        \hline
        \multirow{3}{*}{NL} & DivideMix \cite{li2020dividemix} & 33.00 & 34.72 & 30.67 & 31.34 \\
        & UNICON \cite{karim2022unicon} & 31.86 & 31.12 & 59.47 & 37.06 \\
        & TCL \cite{huang2023twin} & 37.24 & 35.70 & 61.70 & 39.56 \\
        \hline
        \multirow{7}{*}{NL-LT} & MW-Net \cite{shu2019metaweight} & 30.74 & 31.12 & 54.95 & 31.80 \\
        & RoLT \cite{wei2021robust} & 15.78 & 16.90 & 61.23 & 33.48 \\
        & HAR \cite{cao2021har} & 32.60 & 31.30 & 56.84 & 32.34 \\
        & ULC \cite{huang2022uncertainty} & 34.24 & 34.84 & 43.89 & 35.71 \\
        & SFA \cite{li2023sfa} & 36.70 & 35.52 & 63.64 & 40.83 \\
        & TABASCO \cite{lu2023tabasco} & 37.20 & 37.12 & 64.54 & 39.30 \\
        \cline{2-6}
        & DaSC & \bf{40.26} & \bf{39.72} & \bf{70.61} & \bf{44.59} \\
        \Xhline{1.5pt}
    \end{tabular}
    \end{adjustbox}
    \label{tab: imb 100 real-world noise dataset} 
\end{table}

\section{Ablation Study on Hyperparameters}
\begin{figure}[h]
    \centering
        \begin{subfigure}[t]{0.32\columnwidth}
            \centering
            \includegraphics[width = 0.8\columnwidth]{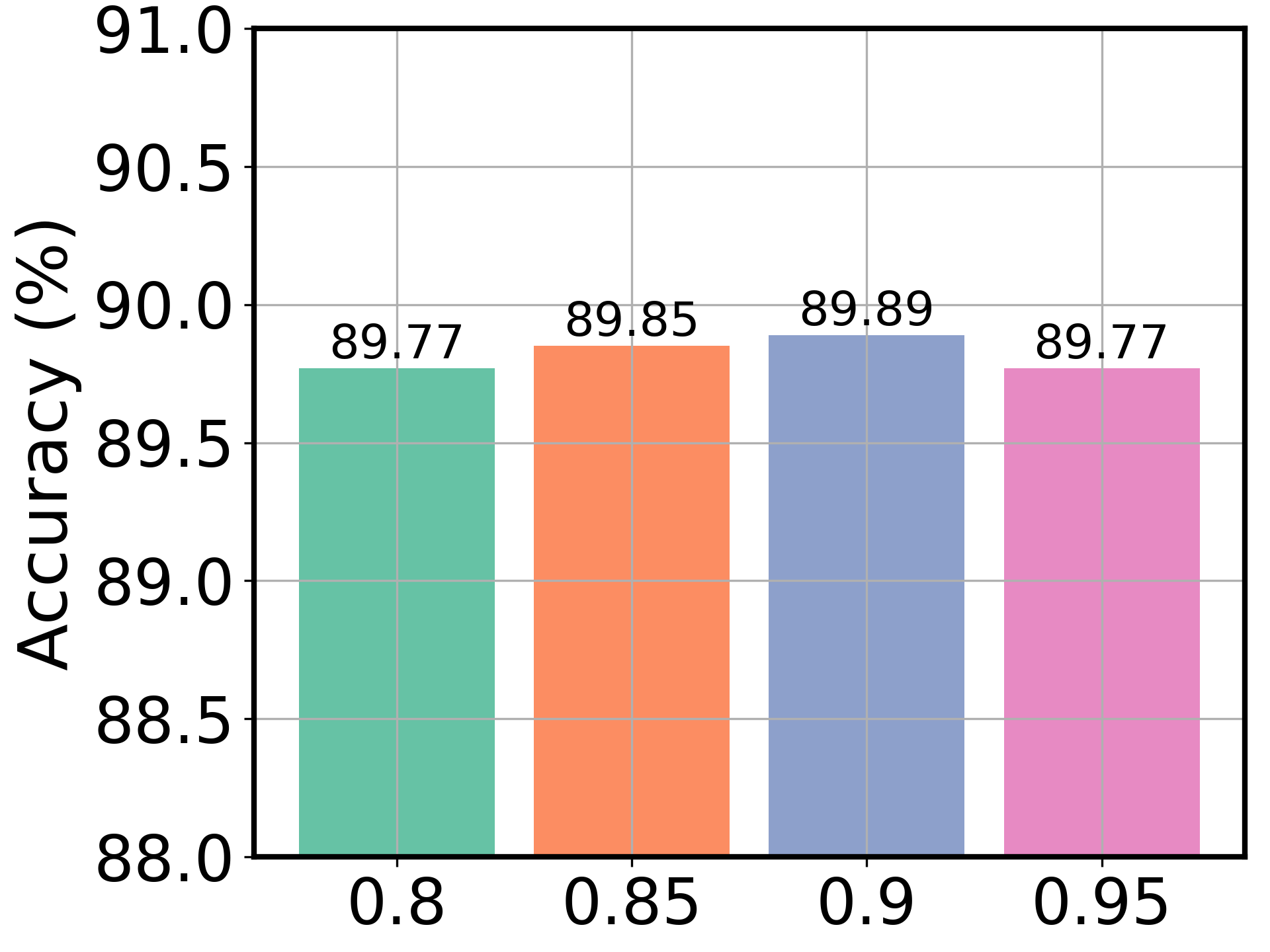}
            \caption[]{Confidence threshold $\tau_c$}
        \end{subfigure}
        \hfill
        \begin{subfigure}[t]{0.32\columnwidth}
            \centering
            \includegraphics[width = 0.8\columnwidth]{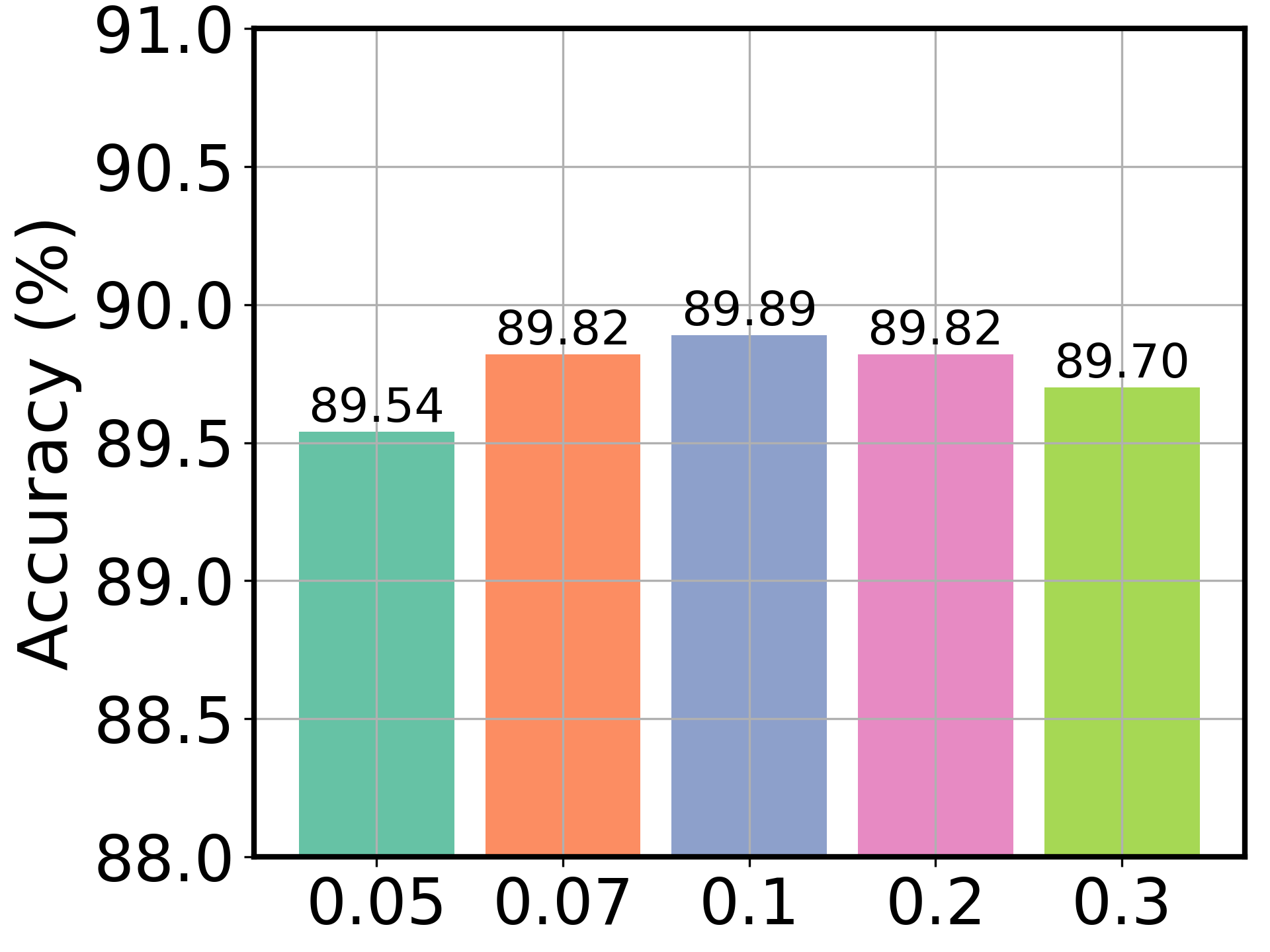}
            \caption[]{Temperature $\tau_T$}
        \end{subfigure}
        \hfill
        \begin{subfigure}[t]{0.32\columnwidth}
            \centering
            \includegraphics[width = 0.8\columnwidth]{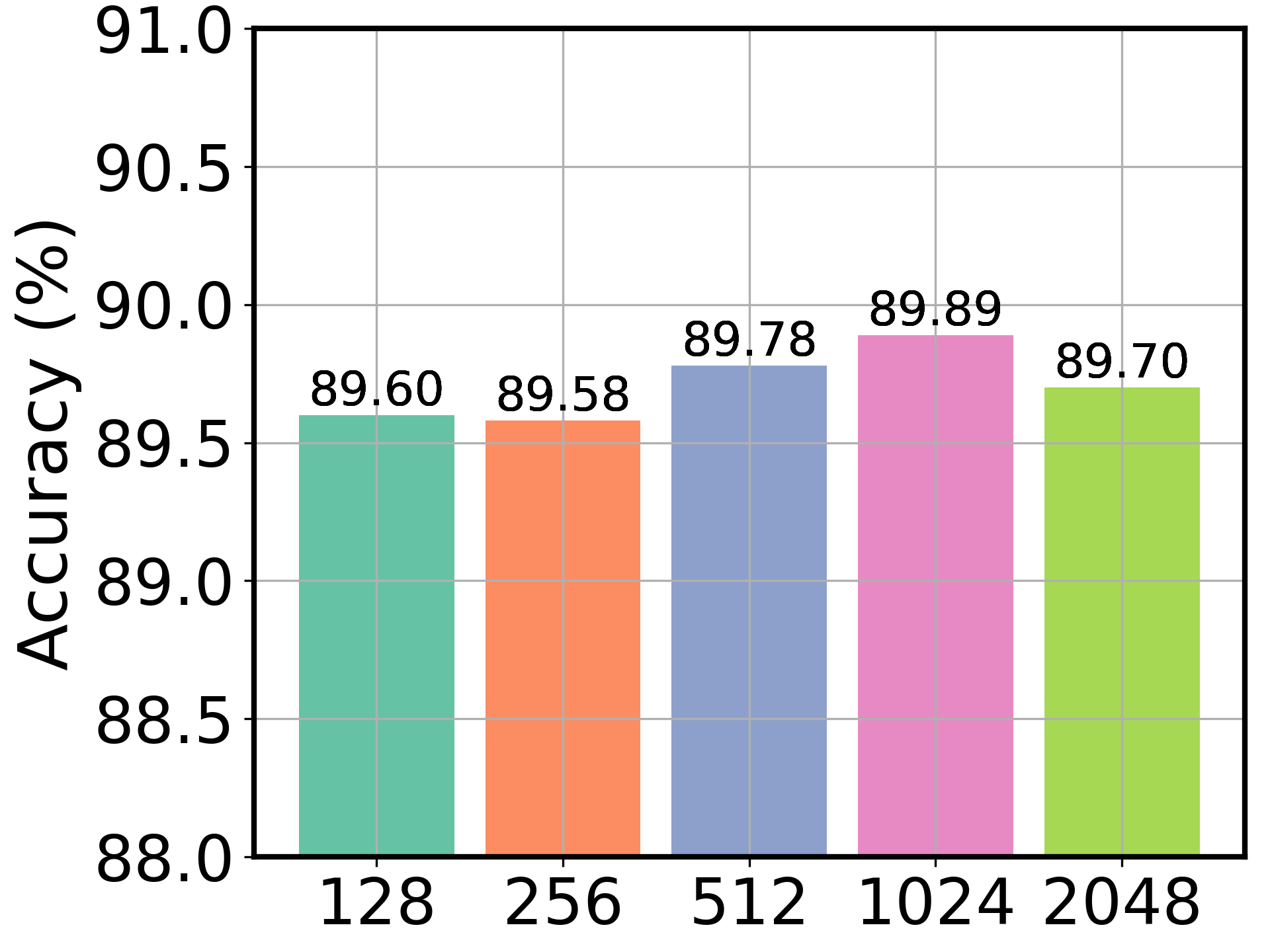}
            \caption[]{Memory bank size $|M|$}
        \end{subfigure}
        \hfill
    \caption[]{Performance comparison with various hyperparameter configurations. The performance was evaluated on long-tailed CIFAR-10 with asymmetric noise of 0.2.}
    \label{fig: hyperparameters asym 0.2}
\end{figure}

\begin{figure}[h]
    \centering
        \begin{subfigure}[]{0.32\columnwidth}
            \centering
            \includegraphics[width = 0.8\columnwidth]{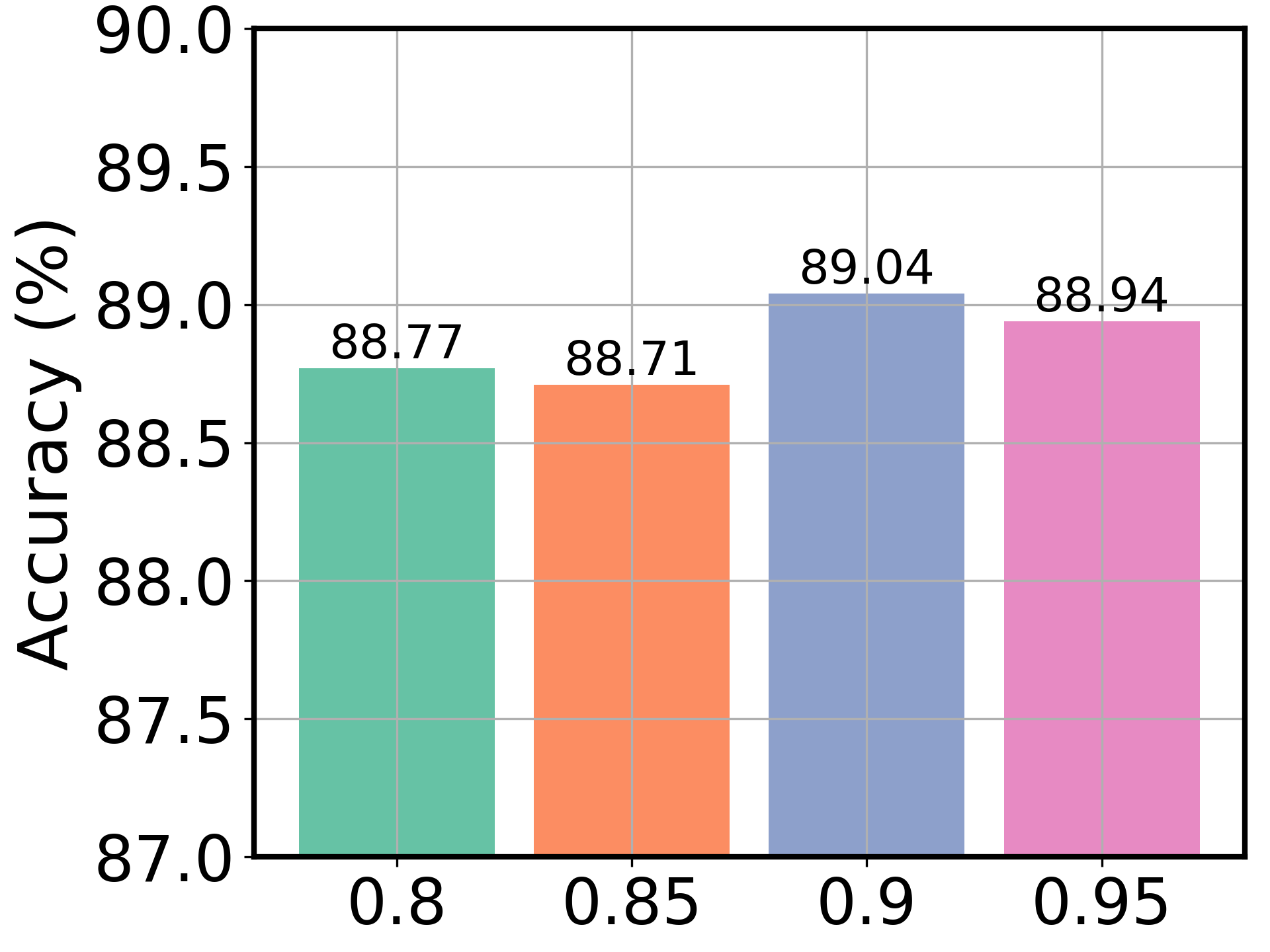}
            \caption[]{Confidence threshold $\tau_c$}
        \end{subfigure}
        \hfill
        \begin{subfigure}[]{0.32\columnwidth}
            \centering
            \includegraphics[width = 0.8\columnwidth]{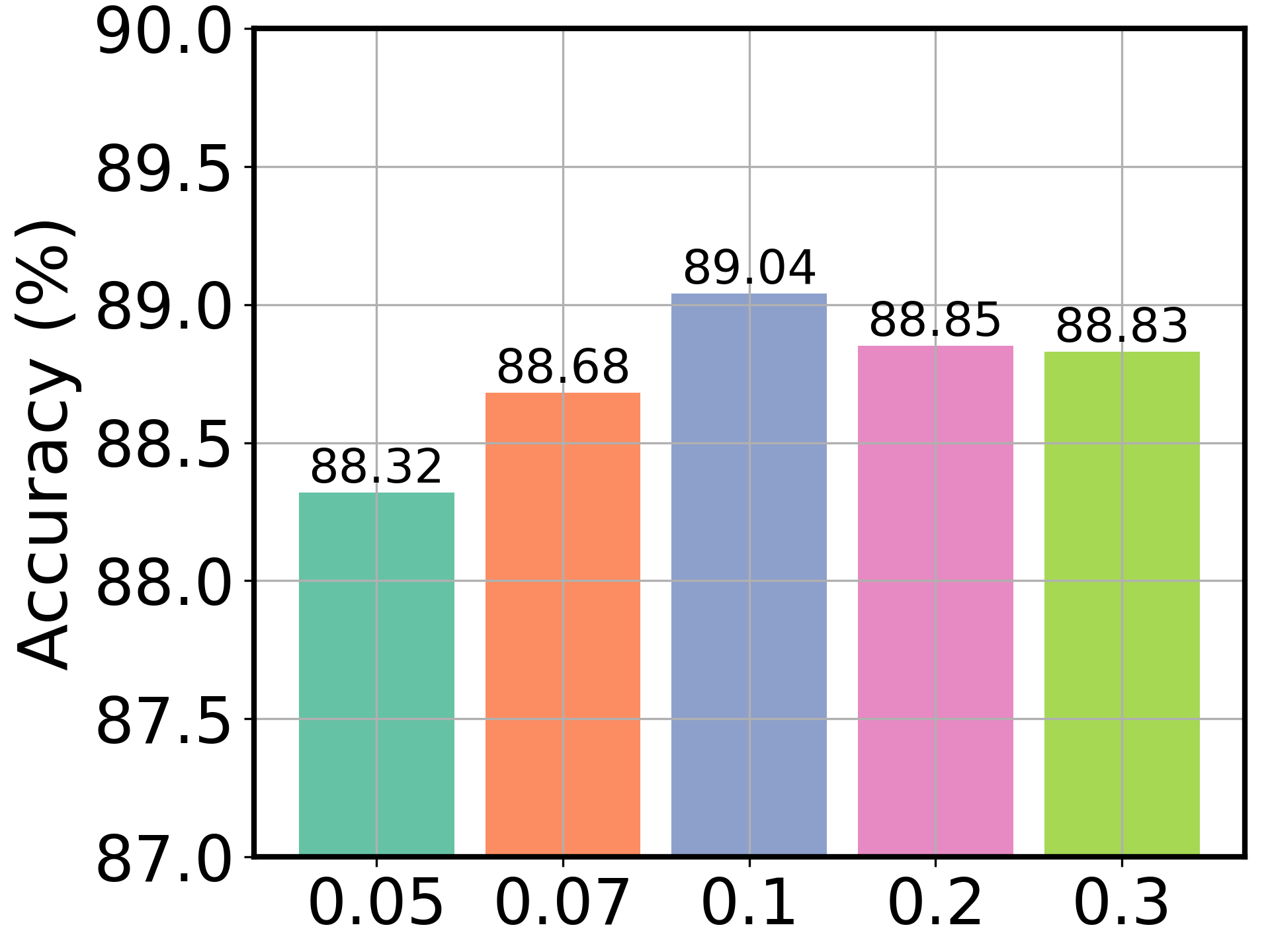}
            \caption[]{Temperature $\tau_T$}
        \end{subfigure}
        \hfill
        \begin{subfigure}[]{0.32\columnwidth}
            \centering
            \includegraphics[width = 0.8\columnwidth]{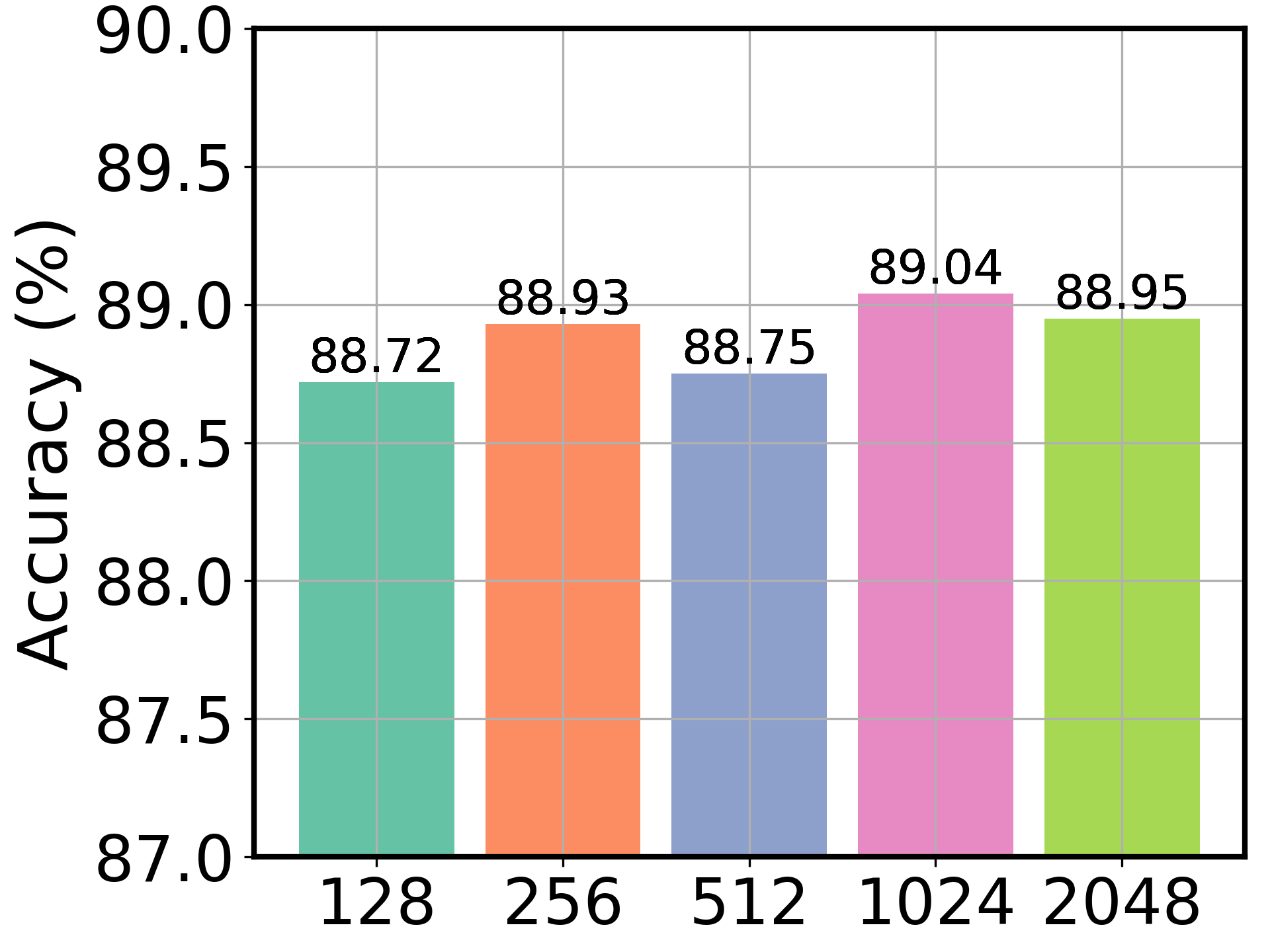}
            \caption[]{Memory bank size $|M|$}
        \end{subfigure}
        \hfill
    \caption[]{Performance comparison with various hyperparameter configurations. The performance was evaluated on long-tailed CIFAR-10 with symmetric noise of 0.4.}
    \label{fig: hyperparameters sym 0.4}
\end{figure}

\begin{figure}[h]
    \centering
    \begin{subfigure}[]{0.243\columnwidth}
        \centering
        \includegraphics[width = 0.8\columnwidth]{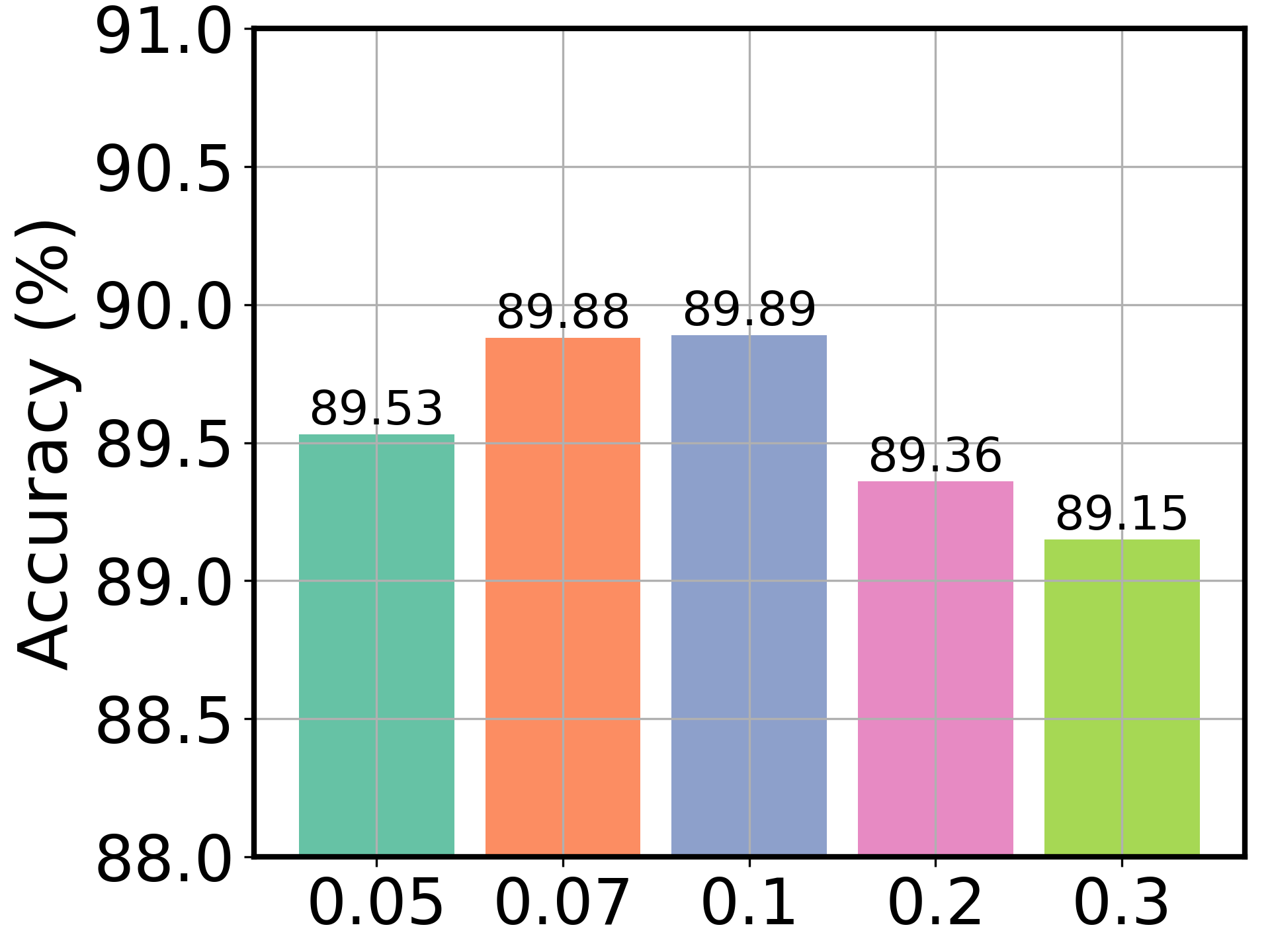}
        \caption[]{Temp. $\tau_s$}
    \end{subfigure}
    \hfill
    \begin{subfigure}[]{0.243\columnwidth}
        \centering
        \includegraphics[width = 0.8\columnwidth]{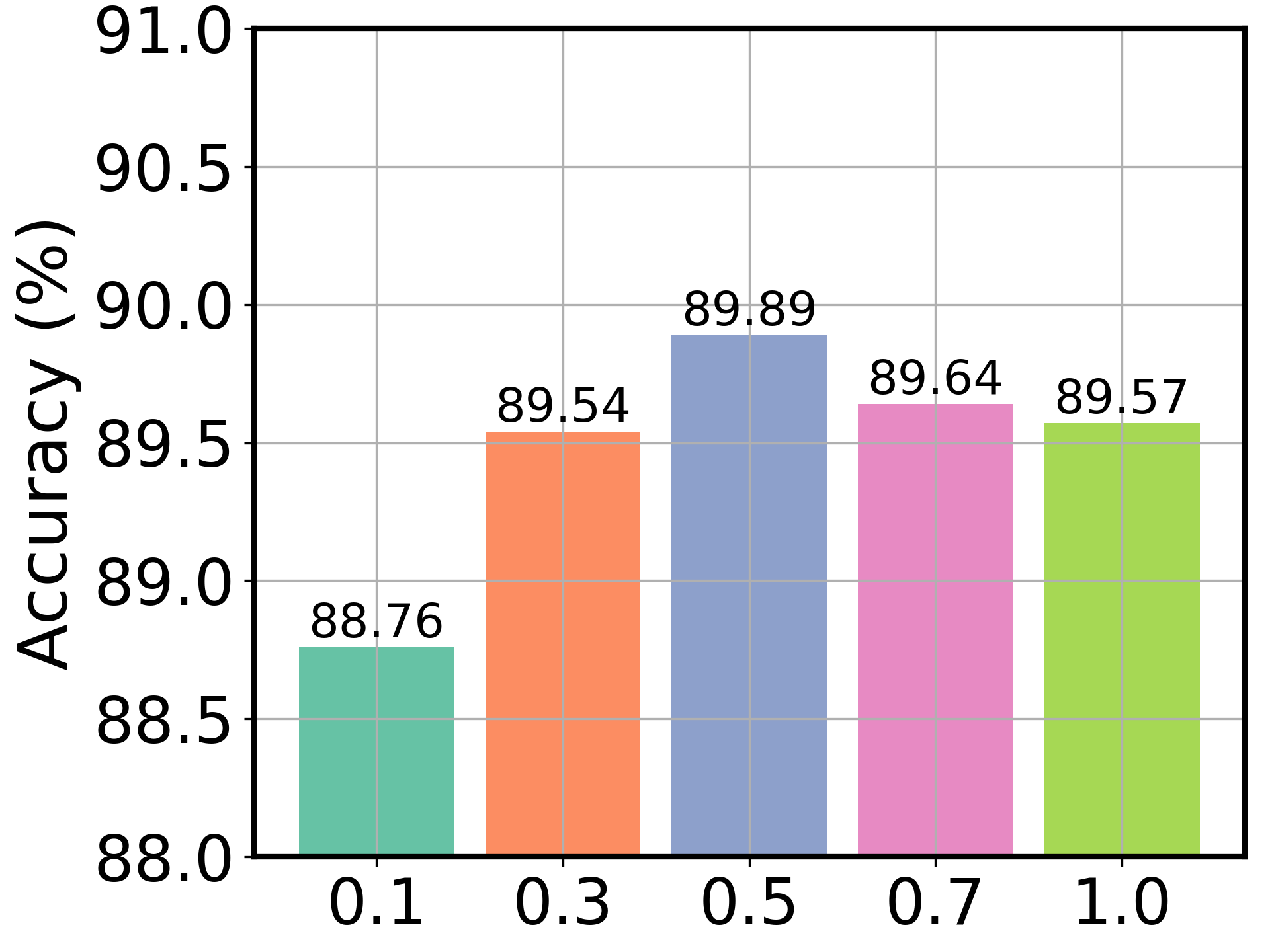}
        \caption[]{Coef. $\lambda_{SBCL}$}
    \end{subfigure}
    \hfill
    \begin{subfigure}[]{0.243\columnwidth}
        \centering
        \includegraphics[width = 0.8\columnwidth]{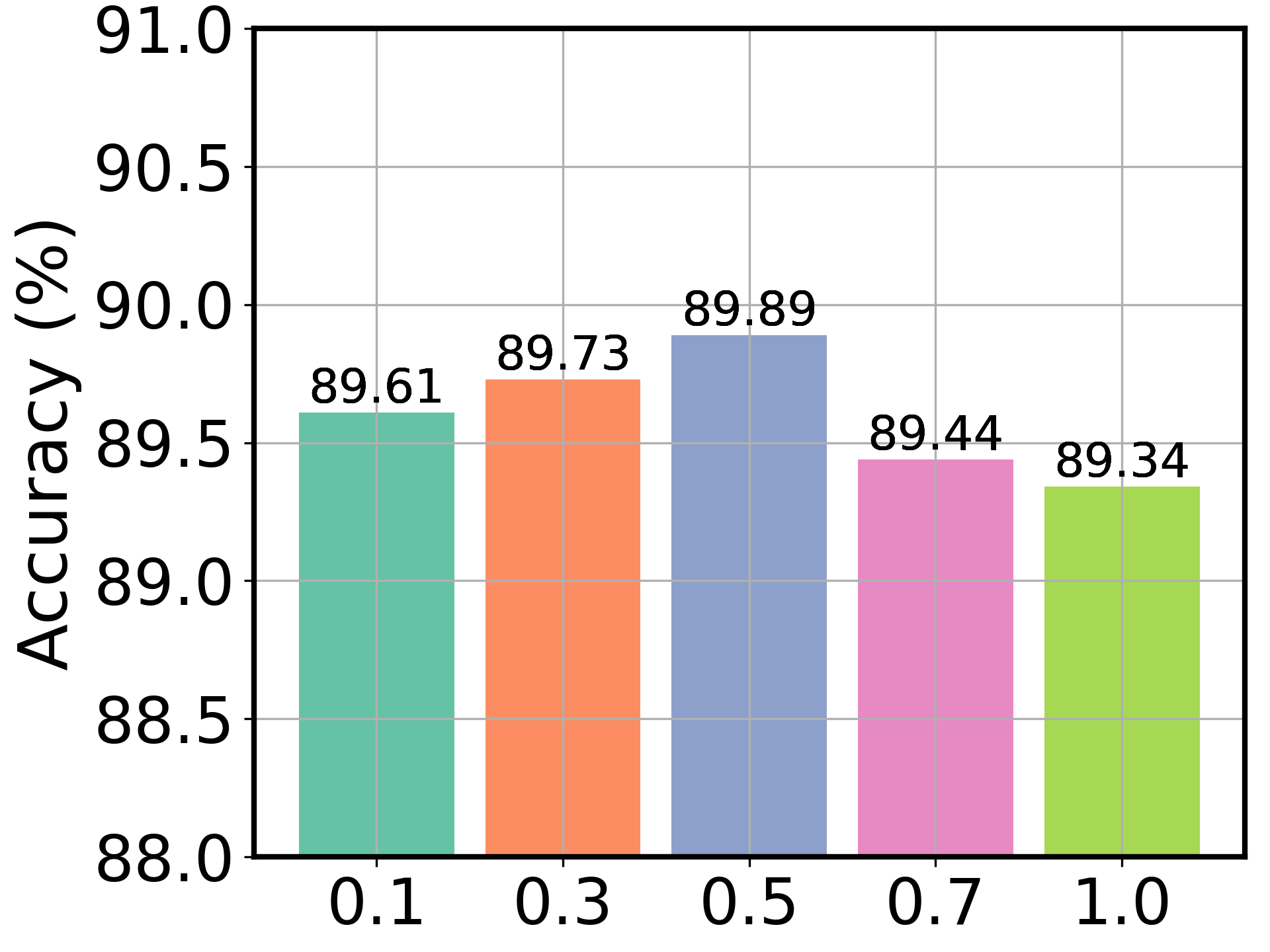}
        \caption[]{Temp. $\tau_m$}
    \end{subfigure}
    \hfill
    \begin{subfigure}[]{0.243\columnwidth}
        \centering
        \includegraphics[width = 0.8\columnwidth]{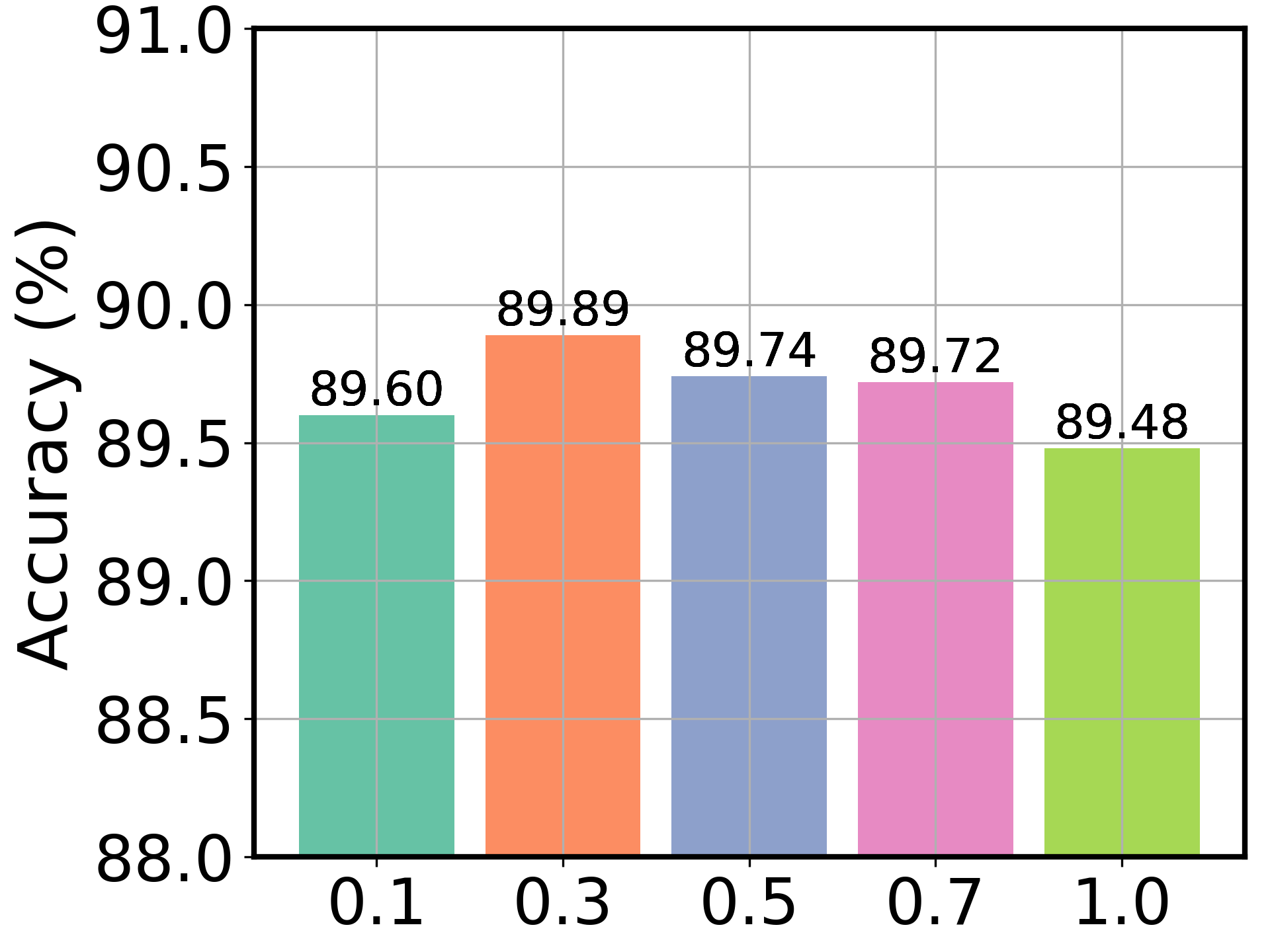}
        \caption[]{Coef. $\lambda_{MIDL}$}
    \end{subfigure}
    \hfill
    \caption[]{Performance comparison with various hyperparameter configurations. The performance was evaluated on long-tailed CIFAR-10 with asymmetric noise of 0.2.}
    \label{fig: hyperparameters_cont asym 0.2}
\end{figure}

\begin{figure}[h]
    \centering
    \begin{subfigure}[]{0.243\columnwidth}
        \centering
        \includegraphics[width = 0.8\columnwidth]{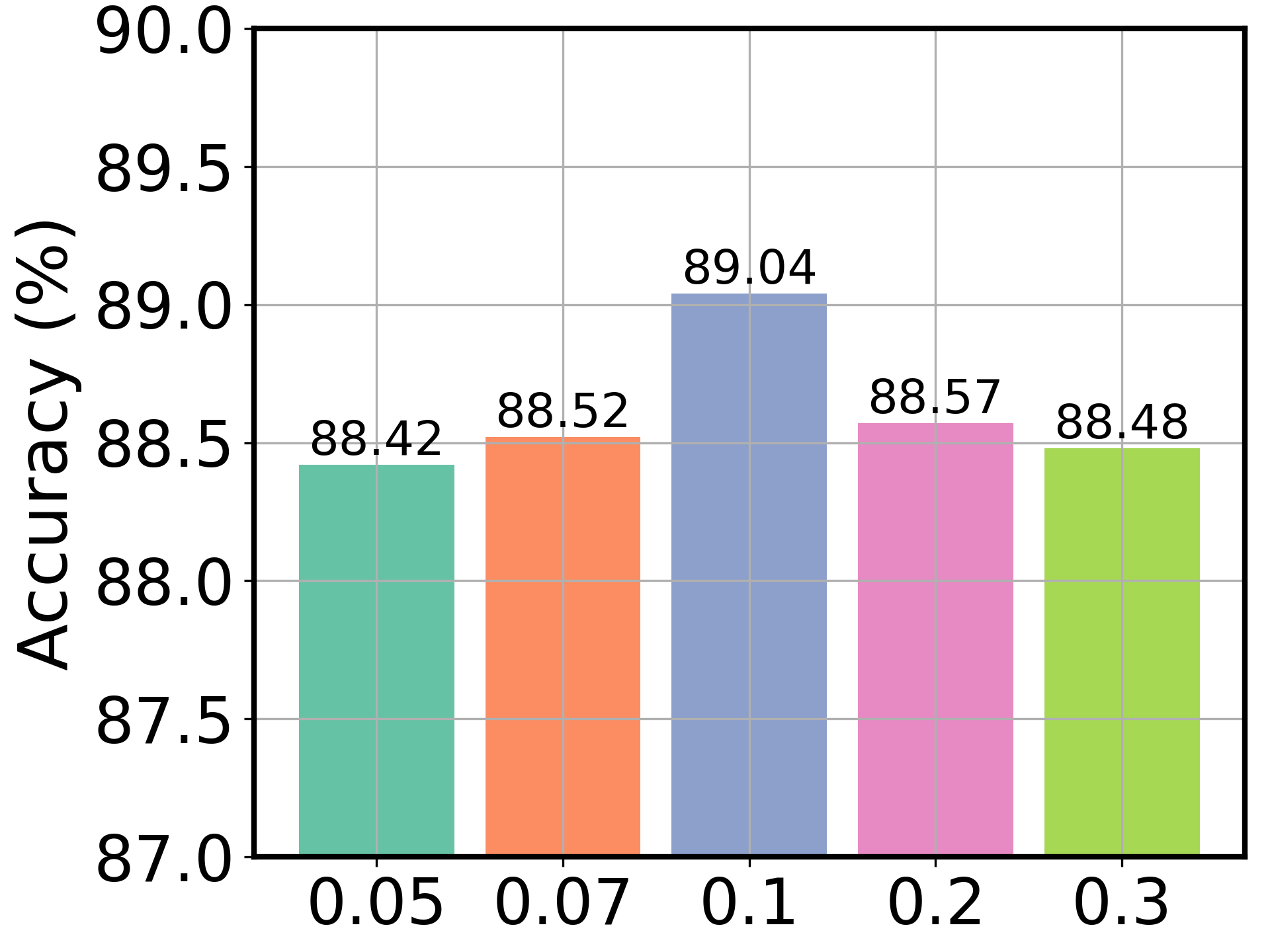}
        \caption[]{Temp. $\tau_s$}
    \end{subfigure}
    \hfill
    \begin{subfigure}[]{0.243\columnwidth}
        \centering
        \includegraphics[width = 0.8\columnwidth]{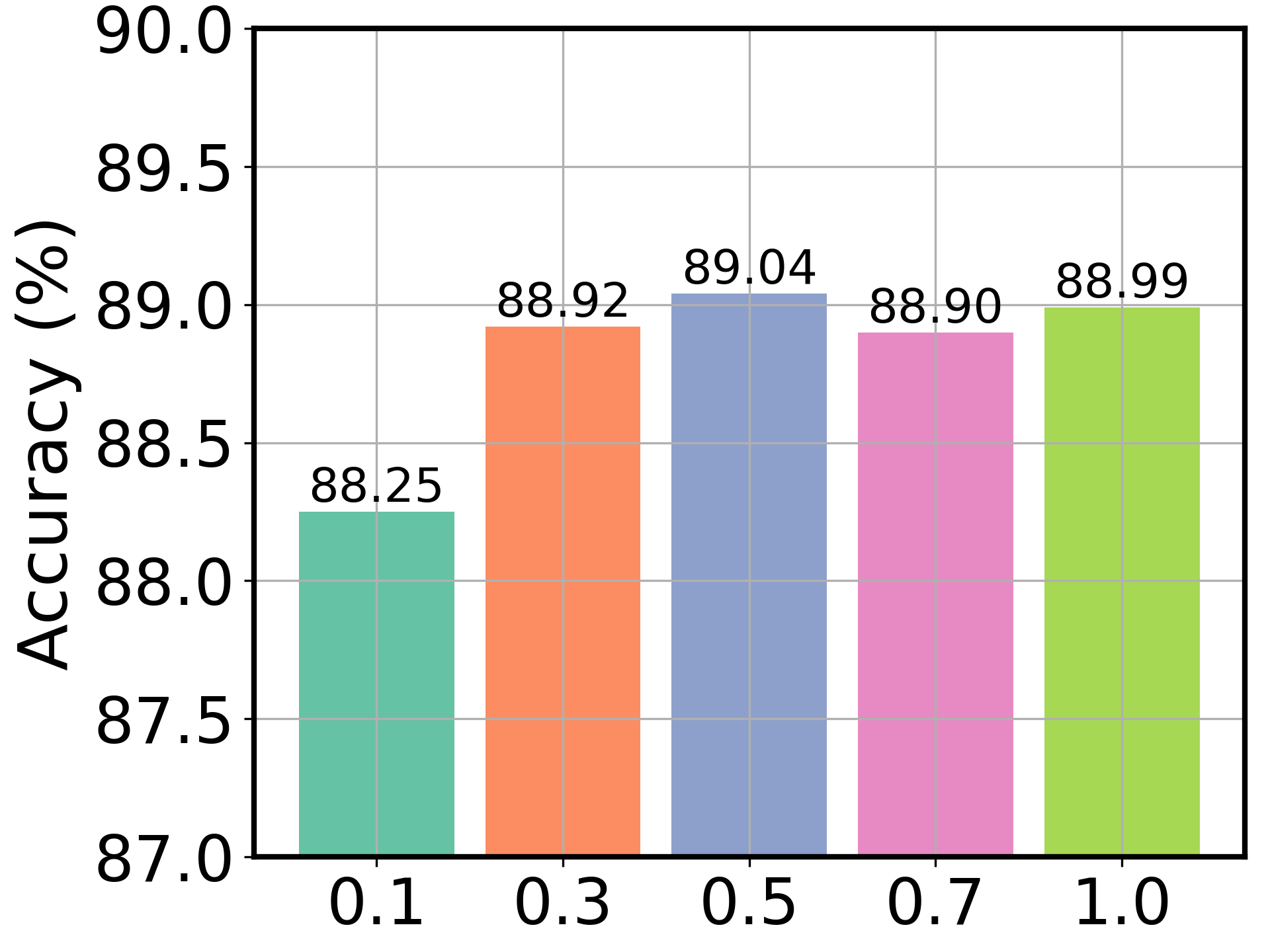}
        \caption[]{Coef. $\lambda_{SBCL}$}
    \end{subfigure}
    \hfill
    \begin{subfigure}[]{0.243\columnwidth}
        \centering
        \includegraphics[width = 0.8\columnwidth]{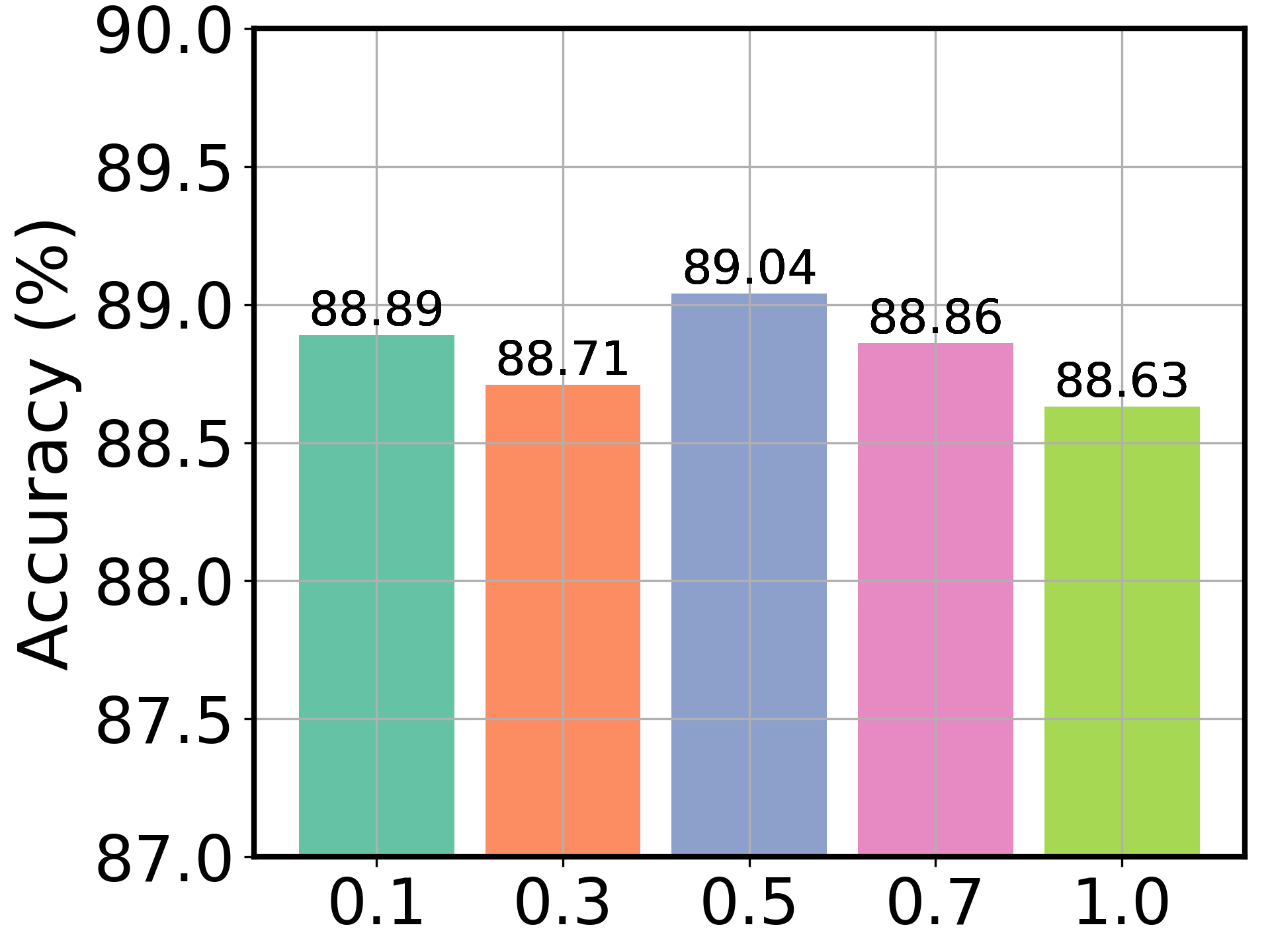}
        \caption[]{Temp. $\tau_m$}
    \end{subfigure}
    \hfill
    \begin{subfigure}[]{0.243\columnwidth}
        \centering
        \includegraphics[width = 0.8\columnwidth]{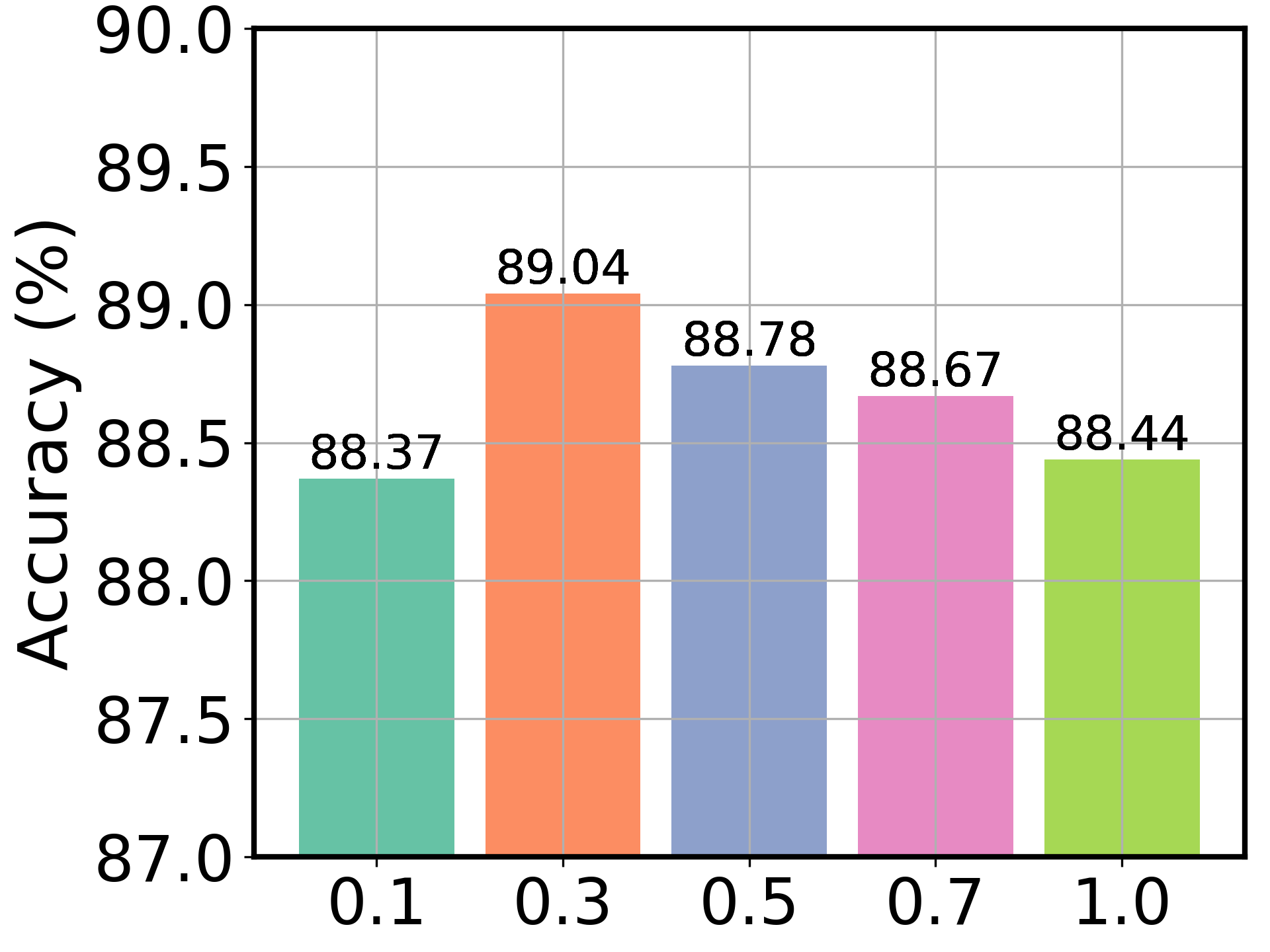}
        \caption[]{Coef. $\lambda_{MIDL}$}
    \end{subfigure}
    \hfill
    \caption[]{Performance comparison with various hyperparameter configurations. The performance was evaluated on long-tailed CIFAR-10 with symmetric noise of 0.4.}
    \label{fig: hyperparameters_cont sym 0.4}
\end{figure}

In Figs. \ref{fig: hyperparameters asym 0.2} and \ref{fig: hyperparameters sym 0.4}, we present experimental results for the confidence threshold $\tau_c$, temperature parameter for temperature scaling $\tau_t$, and memory bank size $|M|$ on the long-tailed CIFAR-10 dataset with asymmetric noise of 0.2 and symmetric noise of 0.4, respectively. Additionally, Figs. \ref{fig: hyperparameters_cont asym 0.2} and \ref{fig: hyperparameters_cont sym 0.4} provide experimental results with other hyperparameters, including the temperature parameter $\tau_s$ for SBCL, the temperature parameter $\tau_m$ for MIDL, the coefficient $\lambda_{SBCL}$ for SBCL, and the coefficient $\lambda_{MIDL}$ for MIDL. These results show that DaSC maintains robust performance across various hyperparameter changes.

\clearpage

\section{Performance of Different Representations and Model Prediction for Noisy Sample Selection.}
Table \ref{tab: representation and prediction} presents the performance of DaSC using different representations and model predictions for detecting correctly labeled samples. For representations, we use either $f(x(i))$ from the backbone network or $z'(i)$ from the MLP projector. For model predictions in DaCC, we employ $\hat{p}^c(i)$ from the conventional classifier or $\hat{p}^b(i)$ from the balanced classifier. 

The results show that the DaSC using $z'(i)$ and $\hat{p}^c(i)$ outperforms all other setups. Using the low-dimensional representation from the MLP projector yields better performance than using the representation from the backbone network. Additionally, predictions from the conventional classifier are more effective than those from the balanced classifier. This is likely due to the early training instability of the balanced classifier, which is trained using an estimate of the data distribution at each epoch, impacting the accurate identification of correctly labeled samples. 

\begin{table}[h]
    \centering
    \caption{Performance of DaSC with different representation and model prediction strategies for class centroid estimation.}
    \begin{adjustbox}{width=0.7\columnwidth}
    \begin{tabular}{c|c|cccc}
        \Xhline{1.5pt}
        \multirow{3}{*}{Representation} & \multirow{3}{*}{Model Prediction} &  \multicolumn{2}{c}{CIFAR-10}& \multicolumn{2}{c}{CIFAR-100}\\
        \cline{3-6}
        & & Sym. & Asym. & Sym. & Asym.\\
        \cline{3-6}
        & & 0.4 & 0.2 & 0.4 & 0.2 \\
        \hline
        $f(x(i))$ & $\hat{p}^c(i)$ & 88.90 & 89.10 & 61.16 & 62.85 \\
        $z'(i)$ & $\hat{p}^b(i)$ & 88.84 & 89.57 & 61.26 & 62.26 \\
        $z'(i)$ & $\hat{p}^c(i)$ &  \bf{89.04} & \bf{89.89} &\bf{61.85} & \bf{63.22} \\
        \Xhline{1.5pt}
    \end{tabular}
    \end{adjustbox}
    \label{tab: representation and prediction}
\end{table}

\clearpage

\section{Ablation Study on Subset $\mathcal{D}^I$}
DaSC leverages samples from a specific subset $\mathcal{D}^I$ rather than directly from the training dataset $\mathcal{D}$. This strategy enhances performance by leveraging a variety of classes to estimate class centroid while filtering out unreliable samples due to noisy labels and long-tailed distributions. As shown in Table \ref{tab: sample subset}, using samples from the subset $\mathcal{D}^I$ achieves better performance than using them directly from the training dataset $\mathcal{D}$. 

\begin{table}[h]
    \centering
    \caption{Performance comparison of using $\mathcal{D}$ versus $\mathcal{D}^I$ used in DaCC.}
    \begin{adjustbox}{width=0.50\columnwidth}
    \begin{tabular}{c|cccc}
        \Xhline{1.5pt}
        \multirow{3}{*}{Sample Set} &  \multicolumn{2}{c}{CIFAR-10}& \multicolumn{2}{c}{CIFAR-100}\\
        \cline{2-5}
         & Sym. & Asym. & Sym. & Asym.\\
        \cline{2-5}
         & 0.4 & 0.2 & 0.4 & 0.2 \\
        \hline
         $\mathcal{D}$ & 87.89 & 89.31 & 61.26 & 62.56 \\
         $\mathcal{D}^I$ & \bf{89.04} & \bf{89.89} &\bf{61.85} & \bf{63.22} \\
        \Xhline{1.5pt}
    \end{tabular}
    \end{adjustbox}
    \label{tab: sample subset}
\end{table}

\section{Effect of Temperature Scaling}
\begin{figure}[h]
    \centering
        \begin{subfigure}[t]{0.48\columnwidth}
            \centering
            \includegraphics[width = 0.9\columnwidth]{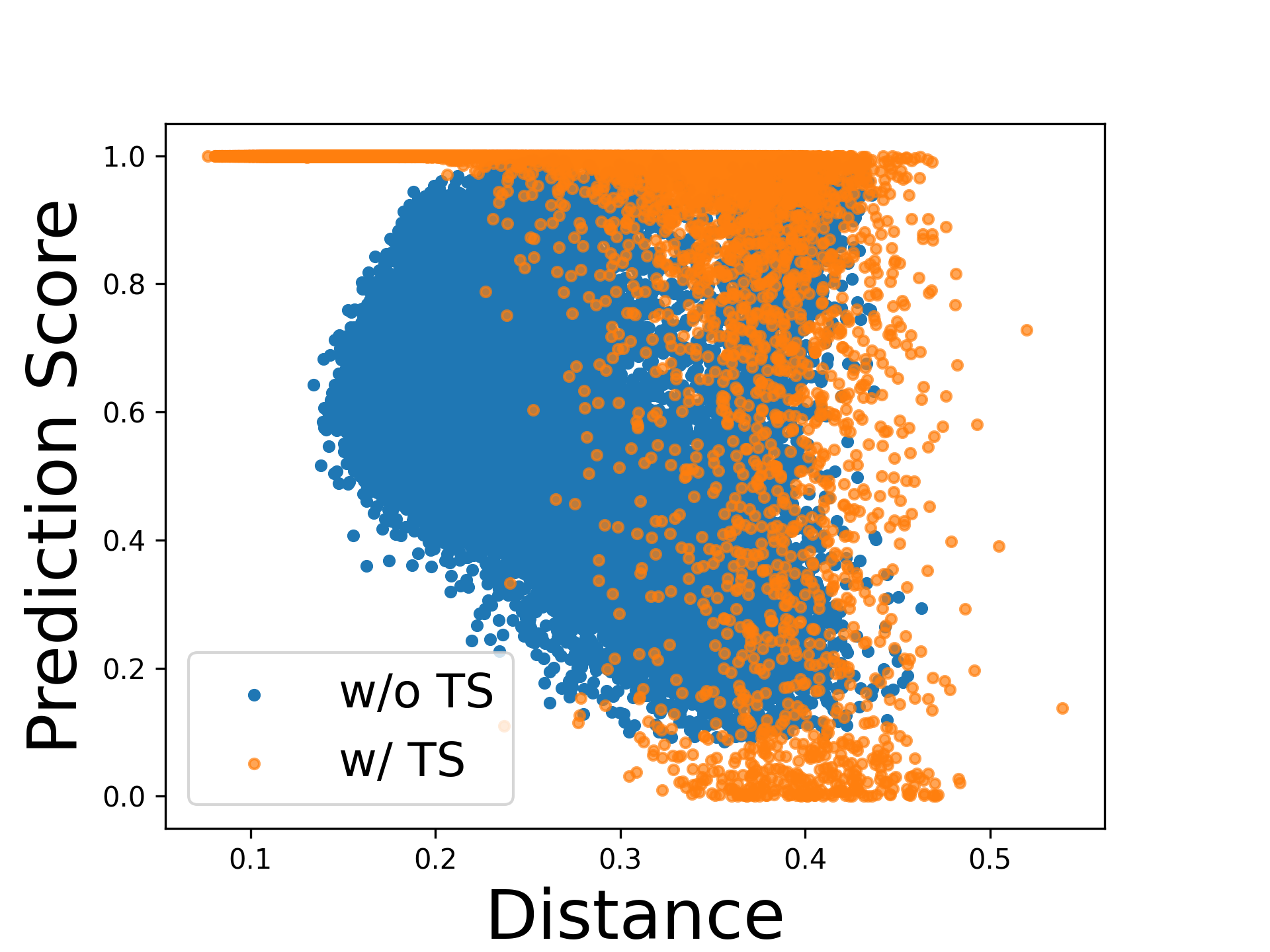}
            \caption[]{Sym. 0.4}
        \end{subfigure}
        \begin{subfigure}[t]{0.48\columnwidth}
            \centering
            \includegraphics[width = 0.9\columnwidth]{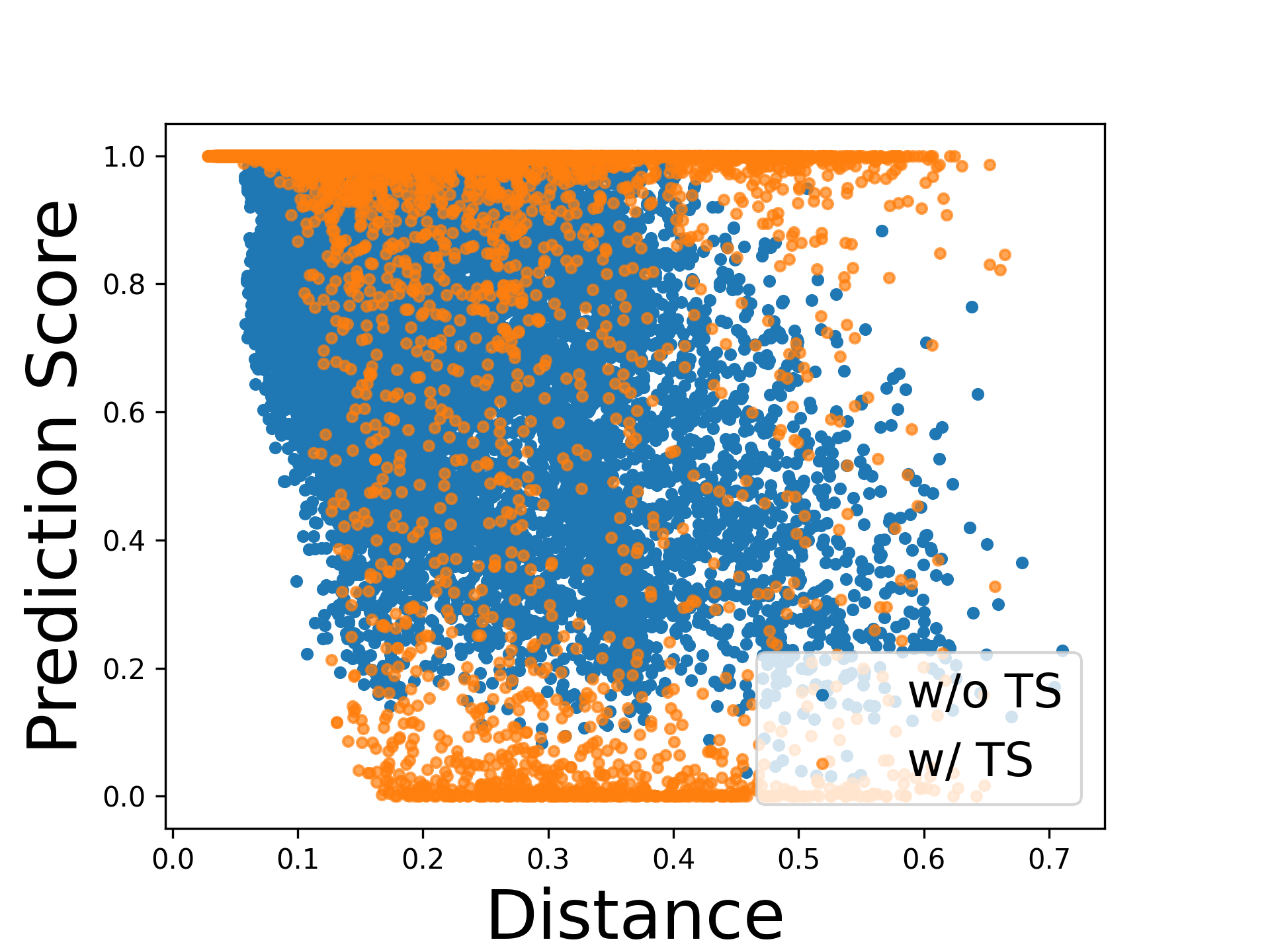}
            \caption[]{Asym. 0.2}
        \end{subfigure}
        \hfill
    \caption[]{Prediction scores of each sample versus the distance to the closest class centroid. `TS' denotes the temperature scaling.}
    \label{fig: pred_val}
\end{figure}

The proposed DaSC employs temperature scaling to mitigate the inherent bias in model predictions. To explore its impact, Fig. \ref{fig: pred_val} presents the prediction score of each sample relative to the distance from the closest class centroid. These results indicate that temperature scaling assigns higher weights to reliable samples closer to the centroid (i.e., higher prediction scores), highlighting their importance, while giving lower weights to unreliable samples farther from the centroids.

\end{document}